\documentclass[journal]{IEEEtran}
\ifCLASSINFOpdf
\else
\fi
\usepackage{times}
\usepackage{epsfig}
\usepackage{graphicx}
\usepackage{amsmath}
\usepackage{amssymb}

\usepackage{url}
\usepackage{booktabs}
\usepackage{multirow}
\usepackage{colortbl}
\usepackage[dvipsnames]{xcolor}
\usepackage[normalem]{ulem}
\usepackage{threeparttable}

\usepackage[breaklinks=true,hidelinks=bookmarks=false,urlcolor=black]{hyperref}
\usepackage[numbers,sort&compress]{natbib}

\definecolor{mygray0}{gray}{0.7}
\newcommand{\clinegrayzero}{\arrayrulecolor{mygray0}\cline{2-12}\arrayrulecolor{black}}
\definecolor{mygray}{gray}{0.9}
\newcommand{\clinegrayone}{\arrayrulecolor{mygray}\cline{2-12}\arrayrulecolor{black}}
\newcommand{\clinegraytwo}{\arrayrulecolor{mygray0}\cline{2-9}\arrayrulecolor{black}}

\newcommand\eg{\emph{e.g.}} 
\newcommand\ie{\emph{i.e.}} 
 
\newcommand\etal{\emph{et al.}}

\begin{document}
\title{Unsupervised Monocular Depth Perception: Focusing on Moving Objects}

\author{Hualie~Jiang,
        Laiyan~Ding,
        Zhenglong~Sun,
        and~Rui~Huang% <-this % stops a space
% <-this % stops a space
\thanks{The authors are with School of Science and Engineering, The Chinese University of Hong Kong, Shenzhen, and also with Shenzhen Institute of Artificial Intelligence and Robotics for Society.} 
}


\markboth{Journal of \LaTeX\ Class Files,~Vol.~14, No.~8, August~XXXX}%
{Jiang \MakeLowercase{\textit{et al.}}: Unsupervised Monocular Depth Perception: Focusing on Moving Objects}



\maketitle

\begin{abstract}
As a flexible passive 3D sensing means, unsupervised learning of depth from monocular videos is becoming an important research topic. It utilizes the photometric errors between the target view and the synthesized views from its adjacent source views as the loss instead of the difference from the ground truth. 
Occlusion and scene dynamics in real-world scenes still adversely affect the learning, despite significant progress made recently. 
In this paper, we show that deliberately manipulating photometric errors can efficiently deal with these difficulties better.  
We first propose an outlier masking technique that considers the occluded or dynamic pixels as statistical outliers in the photometric error map.
With the outlier masking, the network learns the depth of objects that move in the opposite direction to the camera more accurately. 
To the best of our knowledge, such cases have not been seriously considered in the previous works, even though they pose a high risk in applications like autonomous driving. 
We also propose an efficient weighted multi-scale scheme to reduce the artifacts in the predicted depth maps. 
Extensive experiments on the KITTI dataset and additional experiments on the Cityscapes dataset have verified the proposed approach's effectiveness on depth or ego-motion estimation. Furthermore, for the first time, we evaluate the predicted depth on the regions of dynamic objects and static background separately for both supervised and unsupervised methods. The evaluation further verifies the effectiveness of our proposed technical approach and provides some interesting observations that might inspire future research in this direction.

\end{abstract}


% \fi
\IEEEpeerreviewmaketitle

\section{Introduction}
\label{se:intro}

\IEEEPARstart{T}{he} 3D spatial perception for an autonomous system is usually called Simultaneous Localization And Mapping (SLAM), where depth and ego-motion estimation is the core problem.
Although high-precision 3D perception can be achieved via combing the passive camera with active sensors, either ToF~\cite{poggi2019confidence} or LiDAR~\cite{zhao2020fusion}, such an approach is much more expensive and complicated than a purely visual approach. 
Traditional visual SLAM systems primarily rely on multi-view geometry based computer vision algorithms to estimate camera motion and sparse depth~\cite{mur2015orb, engel2014lsd, schoenberger2016sfm}. 
On the other hand, machine learning based methods use massive data to train statistical models to infer depth and camera motion. {The recent development in deep learning has brought success to related tasks, for example, Monocular Depth Estimation (MDE) \cite{
eigen2014depth, wang2015towards, liu2015learning, laina2016deeper, xu2017multi, zeng2017geocuedepth, cao2017estimating, li2018monocular, fu2018deep}, leading to fast and dense 3D reconstruction. }

\begin{figure*}[t]
\begin{center}
\includegraphics[width=0.88\linewidth]{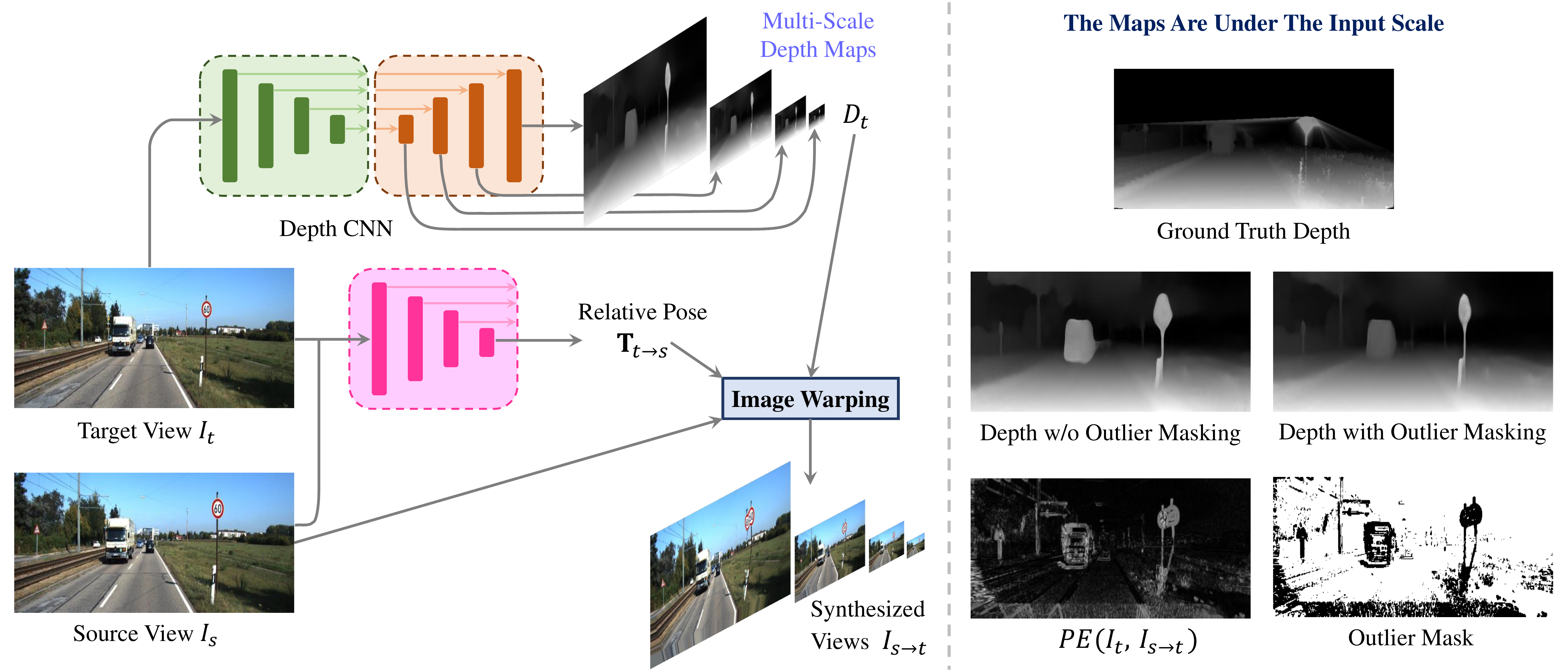}
\end{center}
\vspace{-8pt}
\caption{\textbf{The Unsupervised Learning Pipeline and Effect of Outlier Masking.} {Depth CNN:} A standard fully convolutional U-Net that predicts the multi-scale depth maps for the target image. {Pose CNN:} A standard CNN inputs the target view and one source view and predicts their relative motion. With $D_t$ and $T_{t \to s}$ by the networks, the synthesized image $I_{s \to t}$ from the source view $I_s$ to the target view $I_t$ by differentiable image warping. The photometric errors between $I_t$ and $I_{s \to t}$ can work as the training objective for both the Depth CNN and Pose CNN.
The outlier masking can exclude many invisible and nonstatic pixels, particularly those belonging to contra-moving objects, thus predicting a more accurate depth map. Without outlier masking, the oncoming vehicle is predicted to be closer, and the foreground object significantly dilates.}
\label{fig:framework}
\end{figure*}

Conventional supervised learning requires not only raw data but also ground truth depth, which is also usually obtained by expensive scanning devices with tedious manual work. As video sequences are pervasive on vehicles with mounted cameras~\cite{liu2016vehicle}, more recent unsupervised learning of depth and ego-motion from monocular videos \cite{zhou2017unsupervised, klodt2018supervising, bian2019depth, godard2019digging, wang2020unsupervised, luo2019every, yin2018geonet, zou2018df,  vijayanarasimhan2017sfm, casser2019depth, Gordon_2019_ICCV} has drawn increased attention. 
The unsupervised scheme consists of one deep neural network for predicting the depth map of the target view and another one for estimating the motion between the target view and its temporally adjacent views (see Fig.~\ref{fig:framework}). With the outputs of two networks, the estimated target view can be synthesized by the adjacent source views with image warping. The resulting photometric loss between the target view and synthesized target view works as the supervisory signal for learning monocular depth and ego-motion.

Image reconstruction based supervision assumes that the scene is static and visible from different views. However, this assumption is invalid in the highly dynamic transportation scenario~\cite{yuan2020independent}. 
The image reconstruction is usually corrupted by occlusion and moving objects, and the resulting incorrect supervision harms the network learning.  
Minimizing photometric errors of occluded regions and the moving objects negatively affects their depth estimation. Many methods have tried to deal with occlusion and scene dynamics, producing some improvements. 
For example, the “dark holes” effect caused by the vehicles moving on the same track has been tackled in the latest works \cite{luo2019every, casser2019depth, godard2019digging}.  However, as shown in Fig.~\ref{fig:comparison}, the latest models often significantly underestimate the depth of the oncoming objects, which move towards the camera. This may cause trouble in practice. For instance, if the distance of oncoming vehicles is underestimated in autonomous driving, unnecessary braking or avoiding may be triggered.

{To improve the unsupervised depth learning, we proposed efficient techniques by Delving into Photometric Errors in~\cite{jiang2020dipe}, and we note our proposed approach as DiPE.} In this paper, we extend the previous work by providing additional experiments on the KITTI official single image depth prediction benchmark~\cite{kittidepthserver} and a complete ablation study. {There are also additional experiments on another popular dataset for autonomous driving, Cityscapes~\cite{Cordts2016Cityscapes}, which further demonstrate the effectiveness of the proposed techniques.}

The first proposed technique is the outlier masking, which excludes the occluded and moving regions, especially the oncoming objects. The method is driven by our observation that the photometric errors of occluded and dynamic areas are much larger than that of the static background. {In theory, the visible background usually dominates the scenes and can be easily matched from temporarily adjacent views if depth and ego-motion are approximately predicted. In contrast, the invisible and nonstationary pixels are minor, and the photometric errors of these objects are difficult to minimize. While mainly focusing on the major background, the network could not predict perfectly fake depth for such objects to match adjacent views. Therefore, we exploit the phenomenon by masking these regions as statistical outliers during training.}

Besides, we propose an efficient {weighted multi-scale scheme} for the photometric losses at different scales. The view synthesis in unsupervised depth learning is usually performed in multiple scales by {estimating multi-scale depth maps~\cite{godard2017unsupervised, zhou2017unsupervised, godard2019digging}}. However, the learning will produce texture-copy artifacts due to low-scale errors. {Monodepth2~\cite{godard2019digging}} solves this problem by {upsampling the low scale depth maps to the full resolution} and performing image reconstruction, which increases the training computation. {We instead propose the efficient weighted multi-scale scheme to downweight the artifacts from low scales.} This scheme works nicely with the outlier masking method to produce better depth maps. In fact, the weighted multi-scale scheme is experimentally found to work more compatible with the outlier masking method than the full resolution multi-scale scheme (Tab.~\ref{tab:kitti_eigen_ablation}).

Furthermore, in this paper, we note that the problems caused by the moving objects were often noticed and pointed out by the researchers; however, they were never seriously investigated, especially in a quantitative manner. 
In autonomous driving, the estimation of moving objects is of critical importance. Therefore, we identified five different common motion patterns of the driving scenario in the KITTI dataset and manually labeled them for detailed quantitative investigation. 
The evaluation shows that the proposed outlier masking obtain $0.3\%$, $1.8\%$ and $2.9\%$ improvement on the inlier accuracy $\delta < 1.25$ for the static background, the general moving objects, and the dissimilarly moving vehicles, which indicates that the outlier masking is more effective for moving objects, especially the ones moving differently from the camera. Besides, we also evaluate a supervised method~\cite{fu2018deep} and a stereo image based unsupervised method~\cite{godard2019digging}, showing that they also suffer from the dynamics, though at a lighter level. 
{The gap from the monocular video based unsupervised method to other two types of methods may be the possible room to improve for moving objects. The room is still significant, and more effort should be made for better handling the moving objects.}

In summary, the contributions are:

1) We proposed an outlier masking strategy to mitigate the problematic supervisory signals from the photometric errors of the occlusion and the moving objects;

2) We proposed a weighted multi-scale scheme to reduce the artifacts caused by low-scale photometric error maps; 

3) We provide ground truth for five categories of different moving objects and quantitatively evaluated the performance of our method and many other state-of-the-art methods, both in a conventional manner and focusing on the moving objects.  Our experiments not only demonstrate the effectiveness of our approach, which achieves the new state-of-the-art performance upon existing baselines, but also provide informative observations and discussions on the moving objects.

To facilitate the further investigation, our source code and labeled data will be publicly available at \url{https://github.com/HalleyJiang/DiPE}.

\section{RELATED WORK}
\label{sc:rel-wk}

Traditional geometry based algorithms for estimating the scene structure and camera motion can be categorized into Visual Simultaneous Localization And Mapping (VSLAM)~\cite{mur2015orb, engel2014lsd}, and Structure from Motion (SfM)~\cite{schoenberger2016sfm}. VSLAM is usually applied on a moving camera to track its motion mainly. ORB-SLAM~\cite{mur2015orb} utilizes the matched feature points among frames to track the camera motion and estimate the sparse depth of these points. In contrast, LSD-SLAM~\cite{ engel2014lsd} directly optimizes the photometric error between keyframes on their semi-dense depth maps and relative poses without needing to extract and match feature points. In contrast, SfM can be performed on an arbitrary collection of captured images of the scene, and Multi-View Stereo (MVS)~\cite{schoenberger2016mvs} is usually adopted as post-processing for dense reconstruction. However, traditional approaches require multiple views for estimating depth and may fail in textureless scenes. 
Learning based methods make inferencing depth from a single image possible.

To learn geometry from the image, a natural way is to acquire the annotation of geometry to train the supervised model. Before the rise of deep learning, people usually designed hand-crafted features and adopted graphical models to tackle MDE \cite{saxena20083, liu2010single, ladicky2014pulling, liu2014discrete}, or formulated MDE as a retrieval problem by extracting depth from videos using nonparametric sampling \cite{karsch2014depth}. However, these traditional methods resulted in limited accuracy. 
In recent years, deep Convolutional Neural Networks (CNN) have boosted the performance of MDE, making it possible in real applications. 
One typical approach is using a deep CNN to densely regress the ground truth depth obtained with physical sensors \cite{eigen2014depth, laina2016deeper, zeng2017geocuedepth, jiang2019high}. 
Other approaches are combining deep learning with graphical models \cite{wang2015towards, liu2015learning, xu2017multi} or casting MDE as a dense classification problem \cite{cao2017estimating, li2018monocular, fu2018deep}. 
However, supervised models trained on publicly available datasets with ground truth depth, like the NYU-DepthV2 \cite{silberman2012indoor} or KITTI \cite{geiger2013vision}, usually do not generalize well to real scenarios.

Instead of depending on ground truth, unsupervised learning schemes adopt more easily accessible resources, such as the stereo images \cite{garg2016unsupervised, godard2017unsupervised} or adjacent monocular video frames \cite{zhou2017unsupervised}, to construct the supervisory signal. Garg \textit{et al.} \cite{garg2016unsupervised} proposed the first scheme to train the MDE network with stereo images, where the left image is warped to the right image with the estimated disparity map. The loss is computed as the square photometric error between the wrapped image and the right image. However, the loss is not differentiable, and the first-order Taylor approximation is required. Monodepth \cite{godard2017unsupervised} constructs a differentiable loss by taking the image warping with the bilinear sampling \cite{jaderberg2015spatial}. In addition, the left-right bidirectional consistency, the combination of an $L_1$ and the {S}tructural {SIM}ilarity (SSIM)~\cite{wang2004image} term, and an image-edge-aware disparity smoothness loss borrowed from the stereo matching work \cite{heise2013pm}, together help Monodepth to outperform some supervised methods \cite{eigen2014depth, liu2015learning} on KITTI \cite{geiger2013vision}. {Recently, Tian~etal~\cite{tian2021depth} proposed quadtree-based photometric and depth loss, as well as a cross-layer feature fusion module to further improve depth estimation trained with stereo images.}

The first method training with monocular videos, SfMLearner \cite{zhou2017unsupervised}, adopts an additional Pose CNN to estimate the relative motion between sequential views to make view synthesis attainable. However, the photometric consistency between nearby views is usually unsatisfactory due to occlusion and moving objects. 
To improve this practical framework, many methods have been proposed, which can be mainly classified as masking photometric errors \cite{zhou2017unsupervised, klodt2018supervising, bian2019depth, godard2019digging, wang2020unsupervised}, joint learning with optical flow \cite{wang2020unsupervised, luo2019every, yin2018geonet, zou2018df}, and modeling object motion \cite{vijayanarasimhan2017sfm, casser2019depth, Gordon_2019_ICCV}.

{Most methods introduce auxiliary modules or increase training overhead.} For instance, some masking strategies, the explainability mask \cite{zhou2017unsupervised}, and the uncertainty map \cite{klodt2018supervising} require an extra network to learn. The self-discovered mask by depth inconsistency between the target and source view in SC-SfMLearner \cite{bian2019depth} has to use more GPU memory to make depth inference for the source view during training. 
Joint learning with the optical flow must construct a new network for learning optical flow to explain or compensate for the photometric inconsistency caused by occlusion or scene dynamics. 
Similarly, modeling object motion also requires modules to estimate the segmentation and movement of objects.

Unlike the above methods, DOPlearning~\cite{wang2020unsupervised} uses the overlap mask for occlusion, which is geometrically derived from the view warping process thus is a light-weight design. However, DOPlearning's another mask for occlusion, the blank mask, is derived from the target-to-source view warping. That means that the depth map for every source view has to be predicted in training, significantly increasing training overhead like the self-discovered mask in SC-SfMLearner \cite{bian2019depth}. 

A more straightforward scheme for occlusion is the {minimum reprojection} in Monodepth2~\cite{godard2019digging}, which takes the minimum photometric errors from all source views. It assumes that the pixel in the target view could be occluded in one source view but probably can be seen in at least one source view. As some pixels are ignored, it is also a masking technique. Monodepth2 also uses an {auto-masking} technique for for objects moving at a speed close to the camera speed. This efficient and straightforward masking strategy has been proved effective by Monodepth2, compared with other state-of-the-art methods. However, the oncoming objects have not been noticed and addressed. This paper proposes the outlier masking to handle such objects. Further, our {outlier masking} technique can help the {minimum reprojection} to recover a more accurate boundary for the foreground objects in predicted depth maps. 

To take advantage of both spatial and temporal cues, people used stereo videos for training monocular depth estimation networks \cite{li2018undeepvo, zhan2018unsupervised, yang2018every, luo2019every, godard2019digging, watson2019self}. Although better performance might be achieved this way, in this paper, we will mainly focus on improving the more generalized case, i.e., unsupervised learning from monocular videos.  
{Recently, PackNet-SfM~\cite{guizilini20203d} improved unsupervised depth estimation by using 3D packing modules, but it does not handle the inherent difficulties of the unsupervised framework like ours.}
Besides, we evaluate the performance difference among the supervised, stereo image based unsupervised, and monocular video based unsupervised methods on regions with different motion patterns.

\section{Methodology}
\label{sc:method}
{This section first presents our core analysis and method first, where some notations can be referred to the final basic model.}
\subsection{Analysis of Occlusion and Dynamics}
\begin{figure}[t]
\begin{center}
\includegraphics[width=0.66\linewidth]{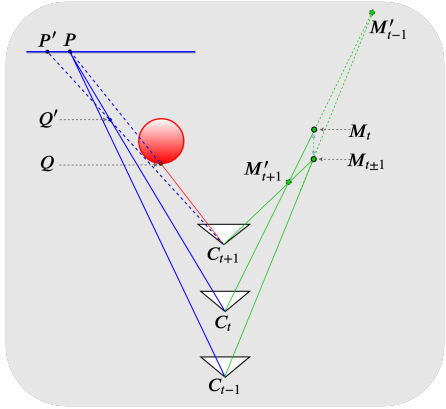}
\end{center}
\vspace{-9pt}
\caption{\textbf{Occlusion and dynamics in unsupervised depth learning.} }
\label{fig:om}
\vspace{-9pt}
\end{figure}

The unsupervised framework above is vulnerable to occlusion and scene dynamics because these factors usually destroy between-view reconstruction.
We illustrate how occlusion and dynamics affect the unsupervised depth learning in Fig.~\ref{fig:om}. 
Suppose that the camera is moving forward from time $t - 1$ to $t + 1$ and $I_t$ is the target view.
The red ball works as an occlusion object for the static blue background. 
In timestamp $t$, the background point $P$ is visible to the camera $C_t$. 
However, in the next timestamp, $t+1$, point $P$ is occluded by the red ball. 
To minimize the photometric error for point $P$, the Depth CNN tends to predict a shorter depth for it, \ie, the depth of $Q'$ and the $P'$ with similar color will be matched with point $P$ instead of the foreground point $Q$. As the region around the foreground object tends to be estimated to be shorter, the foreground object will be blurred; for example, the traffic sign in Fig.~\ref{fig:framework} significantly dilates if disabling the outlier masking.  

For dynamics, we analyze two common motion patterns in the driving scenario. The first one is the co-directional motion to simulate the vehicle moving in the same direction. 
From time $t-1$ to $t$, the point $M$ moves forward like camera $C$. To achieve photometric consistency, the depth of $M$ will be estimated much farther, \ie, that of $M'_{t-1}$. Therefore, such motion produces noticeable `dark holes', which has been identified and tackled in recent works \cite{luo2019every, casser2019depth, godard2019digging}. 
Another motion pattern is the contra-directional motion to simulate the vehicle moving in an inverse direction towards the camera. 
From $t$ to $t+1$, the point $M$ moves in an reverse direction to the camera $C$. 
To optimize the photometric consistency, the depth of $M$ will be estimated shorter, \ie, that of $M'_{t+1}$. 
Probably because the resulting depth inaccuracy is not as significant as that of the co-directional motion, this problem is not identified or seriously treated by previous works. We reveal this case and solve it with the outlier masking.

%-------------------------------------------------------------------------
\subsection{Outlier Masking Technique}
\label{sc:ol-mk}

As has been discussed, some adverse factors, such as occlusion and scene dynamics, result in inappropriate image reconstruction. Therefore some pixels in the photometric error map are invalid, and incorporating them in training can mislead the networks. 
We have observed that most pixels are visible and static, and other occluded and moving regions always produce more significant photometric errors.  This observation inspires the simple but effective outlier masking technique. The outlier mask is automatically determined by the statistical information of photometric errors themselves. 
Technically, we first calculate the mean and standard deviation of pixel photometric errors from all source images for every training sample, 
\begin{eqnarray}
\mu &=& mean{ \{\mathcal{PE}(I_t, I_{s \to t}) | s \in \mathcal{S}\}}, \label{eq:pemean}\\
\sigma &=& std{ \{\mathcal{PE}(I_t, I_{s \to t}) | s \in \mathcal{S}\}}. \label{eq:std}
\end{eqnarray}
Then, for a photometric error map $\mathcal{PE}(I_t, I_{s \to t})$, we compute its outlier mask as,
\begin{equation}
M_s^{ol} = \mu  -l\sigma<\mathcal{PE}(I_t, I_{s \to t}) <  \mu + u\sigma, 
\label{eq:Mol}
\end{equation} 
where $l$ and $u$ are hyper-parameters for the lower and upper thresholds.

\begin{figure}[t]
\begin{center}
\includegraphics[width=0.99\linewidth]{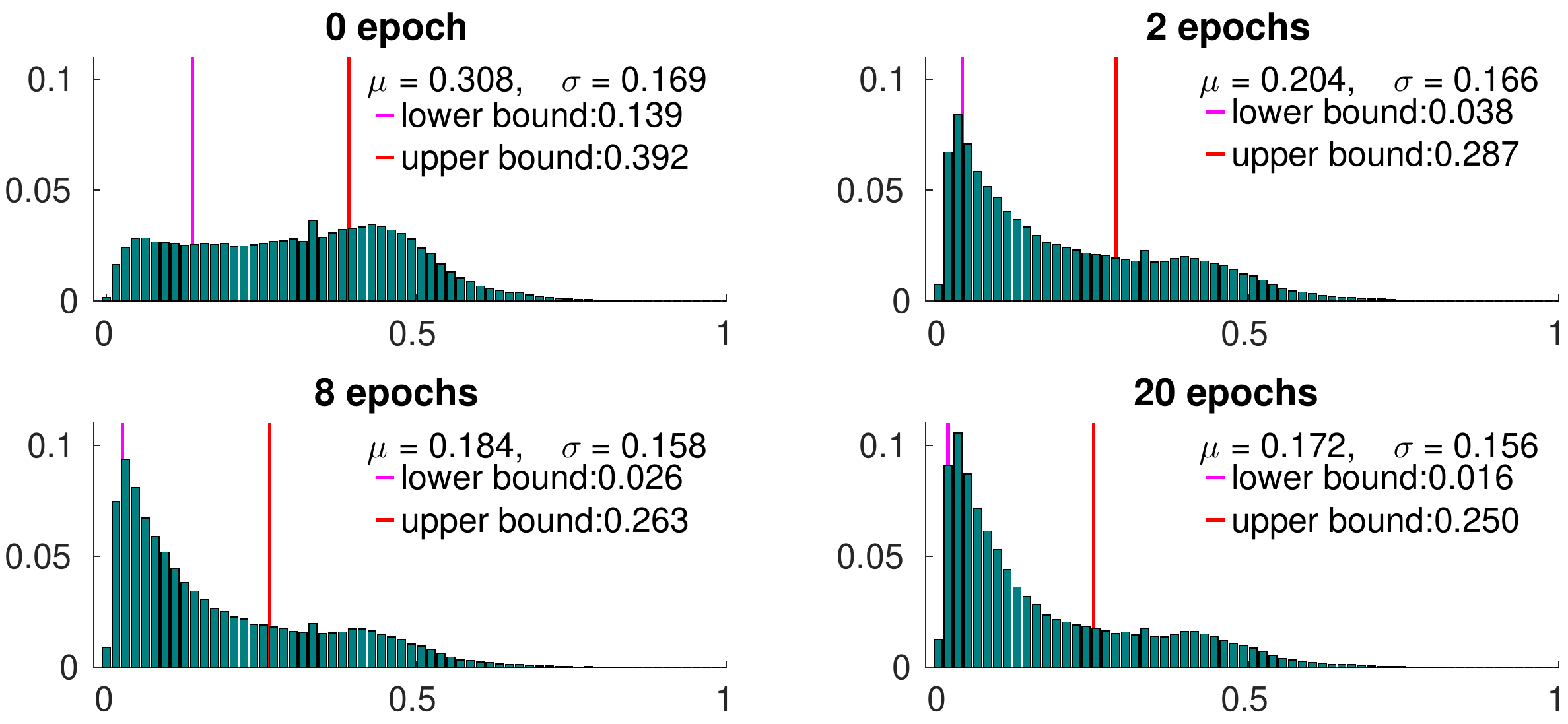}
\end{center}
\vspace{-5pt}
\caption{\textbf{The variation of the photometric error distribution.} The photometric error distribution of a validation sample changes in training precess.  Before training, even with a little long tail, the errors are distributed rather evenly. After two epochs, the majority of errors converge to the lower bound, and a striking long tail forms. Next, the errors under the upper bound continue to decrease and converge, but the errors in the long tail do not vary much. }
\label{fig:pe}
\vspace{-5pt}
\end{figure}

We use this mask to exclude the possible occluded or moving pixels with high photometric errors, as shown in Fig.~\ref{fig:framework}. By the visualization of validation samples in training, we find that 0.5 is a good empirical value for $u$. A lower value may mask out too many stationary pixels while a higher value may not mask out moving objects sufficiently. A lower value can mask out many stationary pixels. Also, $l$ is set as $1$ to mask some pixels with very small photometric errors because these pixels usually belong to homogeneous regions and are not very valuable for network training. This selection for $u$ and $l$ can maintain the principal photometric error for guiding the networks (see Fig.~\ref{fig:pe}).

%-------------------------------------------------------------------------
\subsection{Weighted Multi-Scale Scheme}
\label{sc:w-ms}

{The gradient locality of the bilinear sampler \cite{jaderberg2015spatial} can cause the learning to get stuck in local minima. To avoid this issue, the unsupervised learning models usually predict depth maps in four scales (Fig.~\ref{fig:framework}) and compute multi-scale photometric losses for training. However, it has been pointed out that this scheme tends to produce `holes' in large low-texture regions} {in the intermediate lower resolution depth maps, as well as texture-copy artifacts \cite{godard2019digging}. 
To reduce this phenomenon, Monodepth2 \cite{godard2019digging} adopts a full resolution multi-scale scheme, i.e., to upsample the multi-scale depth maps to the full resolution, operate the image warping using the full-resolution images, and compute photometric losses at the full resolution. }

{However, we find that this full-resolution method significantly raises the computation and GPU memory consumption during training. 
To overcome the phenomenon without increasing training overhead, we propose a weighted multi-scale scheme to devaluate the low-resolution photometric losses and mitigate the disadvantages they cause. 
Explicitly, we define a scale factor $f<1$ to compute the weight for the scale $r$, 
\begin{equation}
w_r =    f^r, 
\label{eq:scale-weight}
\end{equation} 
where $r \in \{0, 1, 2, 3\}$. }

%-------------------------------------------------------------------------
\subsection{Integrated Objective Function}
\label{sec:objective}

Even though the outlier masking technique can exclude the majority of occluded and moving regions, it can not handle all the challenging situations. 
For instance, the outlier masking cannot eliminate the pixels that move out of the image boundary, as illustrated at the bottom of the outlier mask in the left bottom of Fig.~\ref{fig:framework}. 
It is easy to preclude the out-of-box pixels by the principled masking technique \cite{mahjourian2018unsupervised}, which only keeps the pixels reprojected inside the image box of the source images. 
Besides, our outlier masking cannot mask out the objects with a speed close to the camera's speed. These objects are usually falsely estimated to the maximum depth, and the corresponding photometric errors can exactly lie in the statistical inlier region. 
As illustrated in the outlier mask in the left bottom of Fig.~\ref{fig:framework}, the car in the same lane is not clearly masked out. Fortunately, the auto-masking technique in Monodepth2 \cite{godard2019digging} can handle such cases, and after including this masking method, the car in front is not predicted to be very far away. Further, we also find that our outlier masking can collaborate with the minimum reprojection in~\cite{godard2019digging} well to produce more accurate foreground object boundaries in the predicted depth maps. 
Therefore, we build a second baseline model with these three techniques. 

The auto-masking technique precludes regions where the photometric error by reconstruction is larger than the direct photometric differeence between the target image and the source image. The minimum reprojection is also a masking technique, and it only retains the pixels with the minimum photometric error among all source views. 
We denote the masks of these three masking methods for the photometric error map $\mathcal{PE}(I_t, I_{s \to t})$ as $M_s^{p}$ (principled masking), $M_s^{a}$ (auto-masking) and $M_s^{mr}$ (minimum reprojection), 

\begin{eqnarray}
M_s^{p} &=& [p_s  \text{ within image box}], \label{eq:Mp} \\
M_s^{a} &=& \mathcal{PE}(I_t, I_{s \to t}) < \mathcal{PE}(I_t, I_s), \label{eq:Ma} \\
M_s^{mr} &=& \mathcal{PE}(I_t, I_{s \to t}) \leq  \min_s{\mathcal{PE}(I_t, I_{s \to t})}, \label{eq:Mmr}
\end{eqnarray}
where $p_s$ is calculated by Eqn.~\ref{eq:proj}. 
Then we can compute the final mask for the photometric error map $\mathcal{PE}(I_t, I_{s \to t})$ by combining the four masks, 
\begin{equation}
M =  M_s^{ol} \bullet M_s^{p} \bullet M_s^{a} \bullet M_s^{mr} ,
\label{eq:Mask}
\end{equation} 
where $\bullet$ denotes the element-wise logical conjunction. 
Finally, the overall objective function is computed by,
\begin{equation}
L = \eta \sum_{r} f^r \sum_{s} \frac{M_sP_s}{\#\{M_s=1\}} +\lambda\sum_{r} e^rL_{es}^r, 
\label{eq:loss}
\end{equation} 
where we express $\mathcal{PE}(I_t, I_{s \to t})$ as $P_s$, $\eta$, and $\lambda$ are weights to balance the two types of losses, and $e$ is a weighted factor for the edge-aware smoothness loss term from different scales.

\subsection{The Basic Pipeline}
\label{sc:prel}

The monocular unsupervised learning pipeline is shown in Fig.~\ref{fig:framework}.
A training sample contains the target frame $I_t$ at time $t$ and some source frames $I_s$ at nearby times, $s \in \mathcal{S}$. Conventionally,  $\mathcal{S} = \{t-1, t+1\}$ or $\{t-2, t-1, t+1, t+2\}$. Suppose that ${\mathbf K}$ is the shared intrinsic matrix of the camera. With the predicted depth $D$ of $I_t$ by the Depth CNN and the rigid motion ${\mathbf T}_{t \to s}$ between $I_t$ and $I_s$ by the Pose CNN, we can synthesize from the target view $I_t$ from the source view $I_s$ by,
\begin{equation}
\quad I_{s \to t} = I_{s}\langle proj(D, {\mathbf T}_{t \to s}, {\mathbf K}) \rangle, 
\label{eq:synth}
\end{equation}
where $\big\langle\big\rangle$ is the differentiable bilinear sampling operator \cite{jaderberg2015spatial} and $proj()$ is the operation projecting the pixel $p_t$ in the target image to the point $p_s$ in the source image, 
\begin{equation}
p_s \simeq {\mathbf K}{\mathbf T}_{t \to s}D(p_t){\mathbf K}^{-1}p_t,
\label{eq:proj}
\end{equation}
where $p_t$ and $p_s$ are expressed in homogeneous coordinates.

We we adopt the conventional combination of $L_1$ norm and SSIM by \cite{godard2017unsupervised} to compute the photometric errors, 
\begin{equation}
\mathcal{PE}(I_a, I_b) = \alpha \frac{1 - \mathrm{SSIM}(I_a, I_b)}{2} + (1-\alpha) \|I_a - I_b\|_1,
\end{equation}
where $\alpha$ is usually set as $0.85$.

\begin{figure*}[!ht]
\centering
\resizebox{\textwidth}{!}{
\newcommand{\turnheightnew}{0.176\columnwidth}

\centering

\renewcommand{\arraystretch}{0.5}
\begin{tabular}{@{\hskip 1mm}c@{\hskip 1mm}c@{\hskip 1mm}c@{\hskip 1mm}c@{\hskip 1mm}c@{}}

{\rotatebox{90}{\hspace{5mm}\scriptsize Input}} &
\includegraphics[height=\turnheightnew]{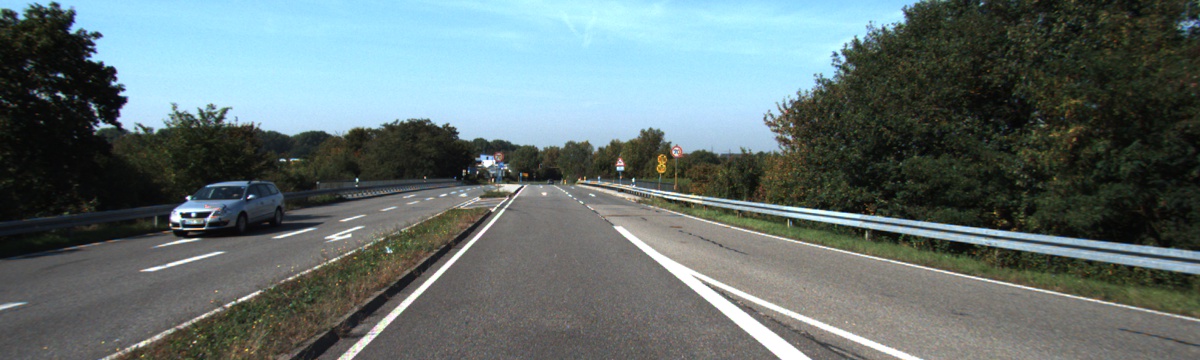} &
\includegraphics[height=\turnheightnew]{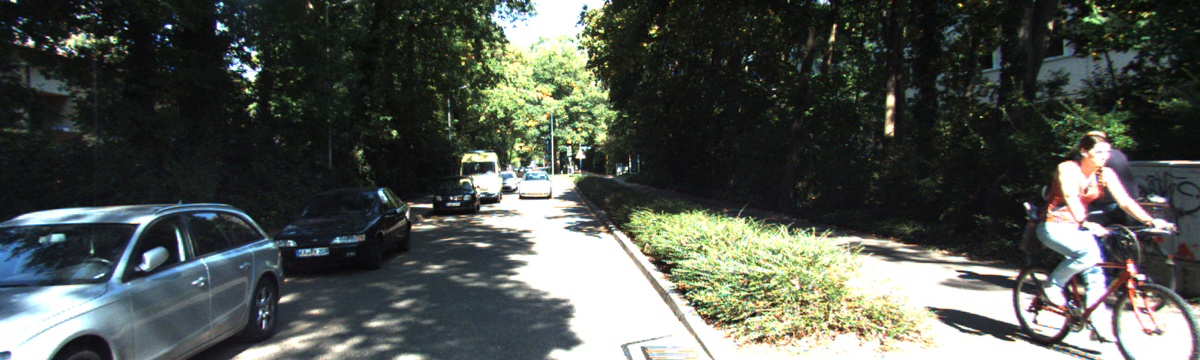} &
\includegraphics[height=\turnheightnew]{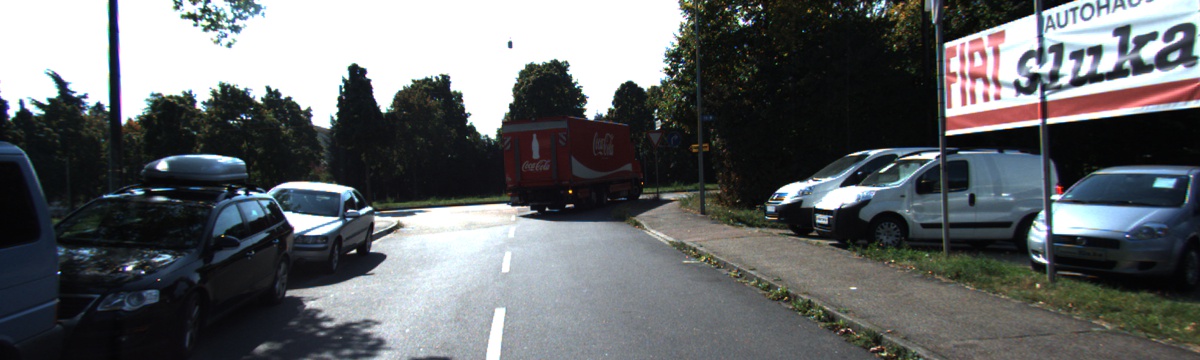} &
\includegraphics[height=\turnheightnew]{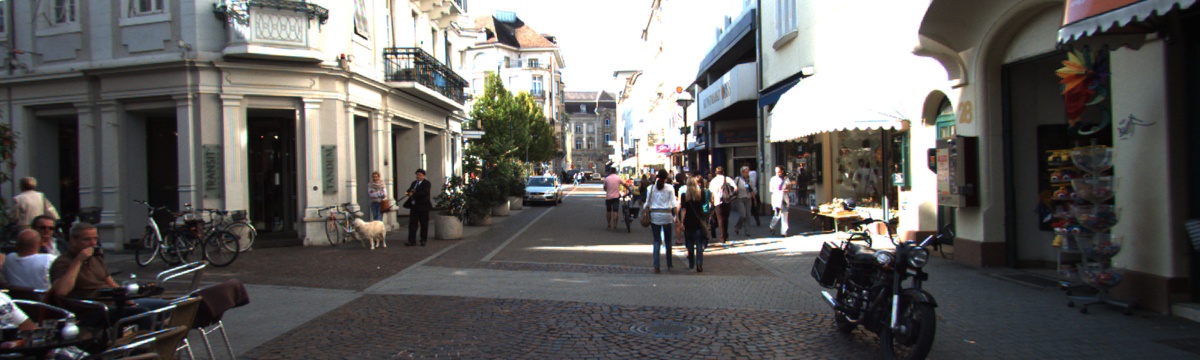}\\

{\rotatebox{90}{\hspace{0.5mm}\scriptsize
{Ground Truth}}} &
\includegraphics[height=\turnheightnew]{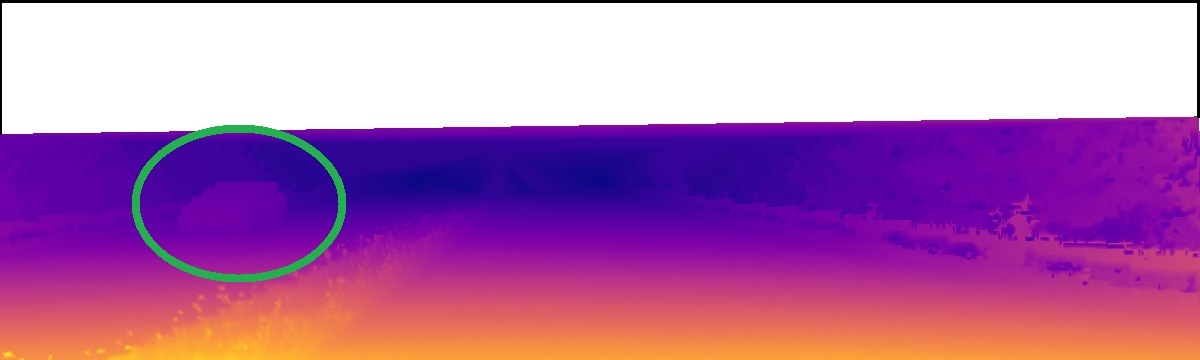} &
\includegraphics[height=\turnheightnew]{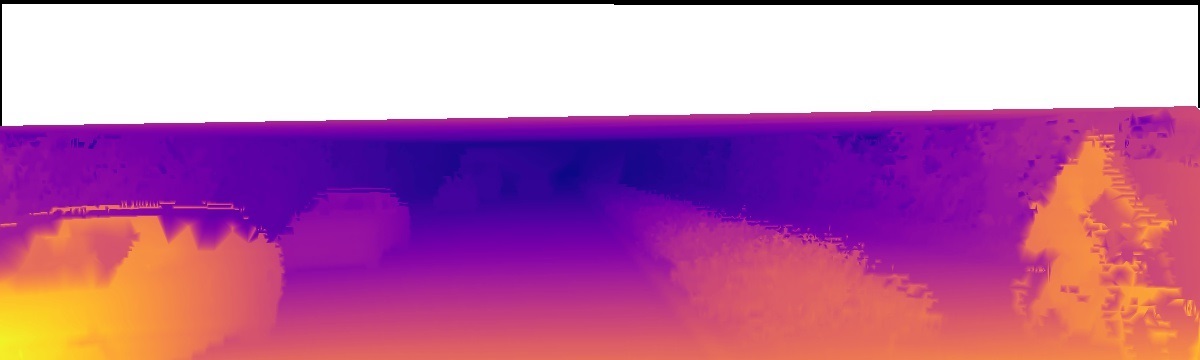} &
\includegraphics[height=\turnheightnew]{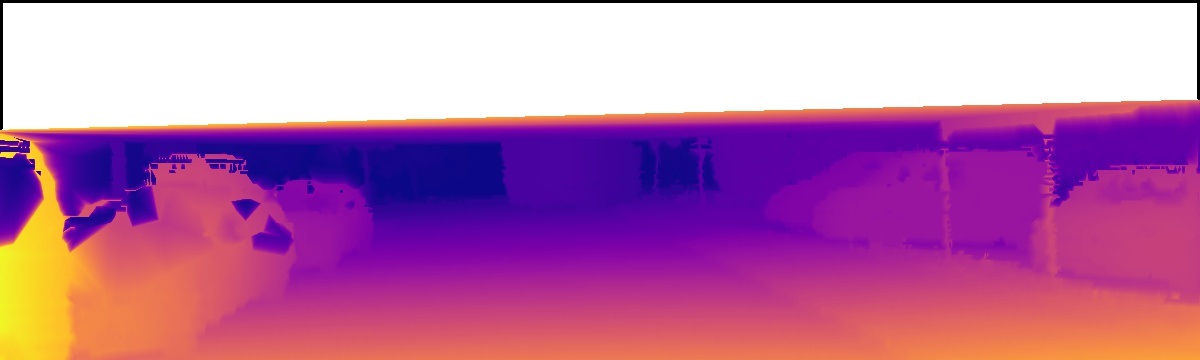} &
\includegraphics[height=\turnheightnew]{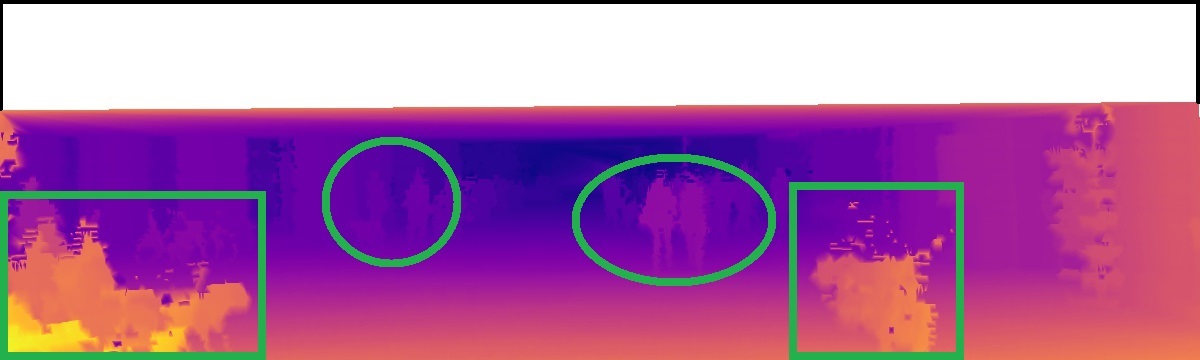}\\

{\rotatebox{90}{\hspace{-0.1mm}\scriptsize
{SfMLearner\cite{zhou2017unsupervised}}}} &
\includegraphics[height=\turnheightnew]{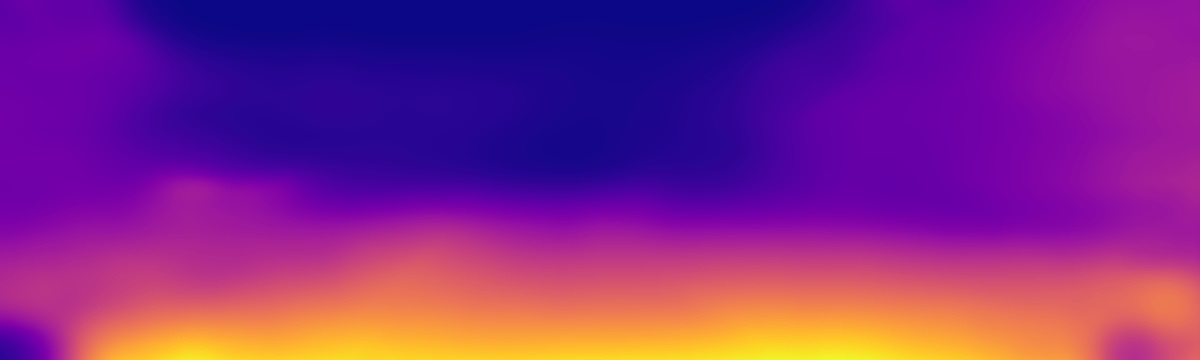} &
\includegraphics[height=\turnheightnew]{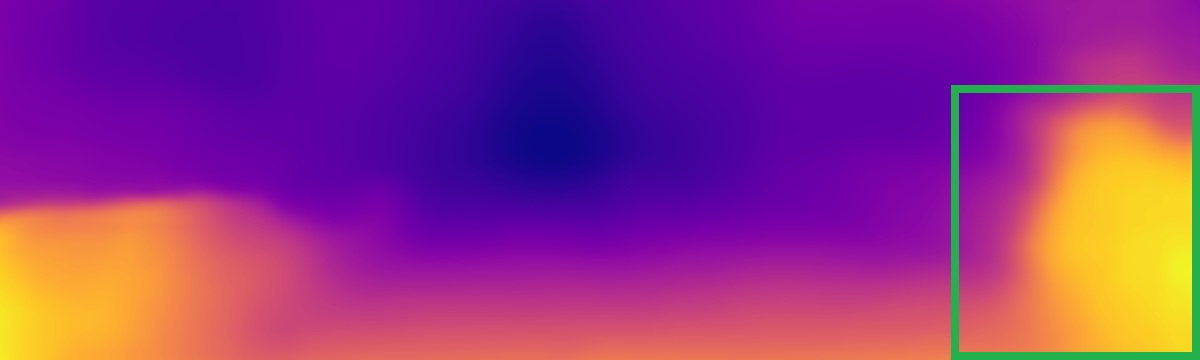} &
\includegraphics[height=\turnheightnew]{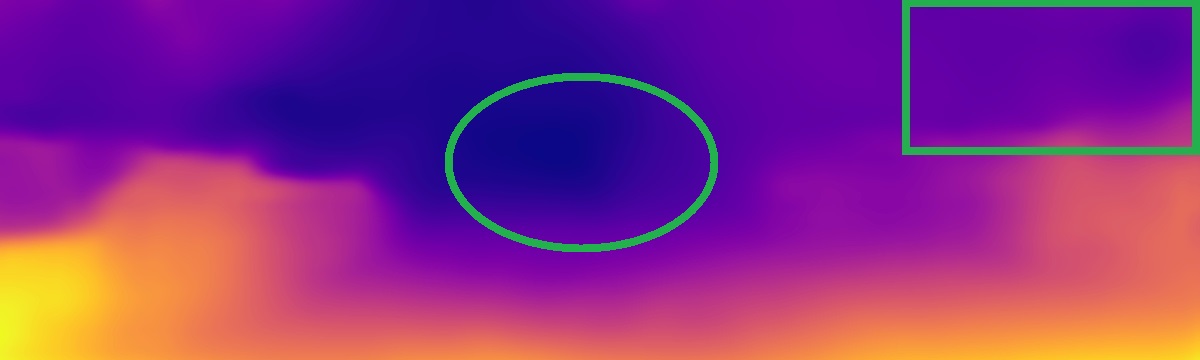} &
\includegraphics[height=\turnheightnew]{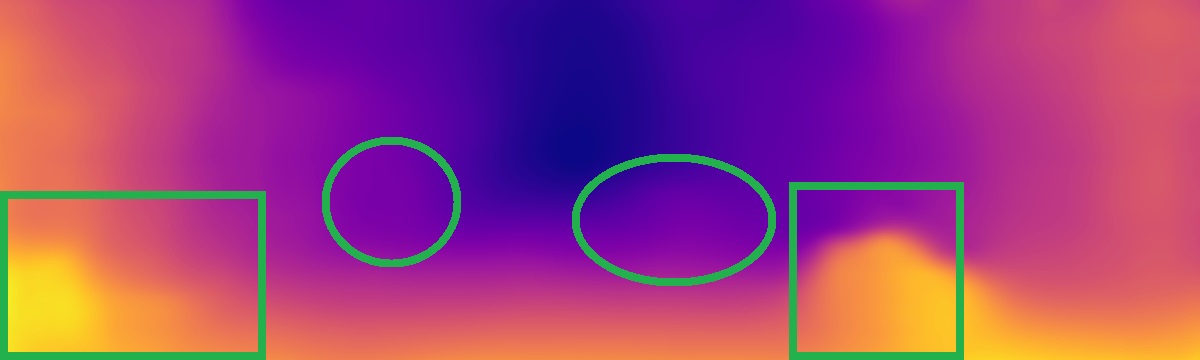}\\

{\rotatebox{90}{\hspace{0.5mm}\scriptsize
{ GeoNet\cite{yin2018geonet}}}} &
\includegraphics[height=\turnheightnew]{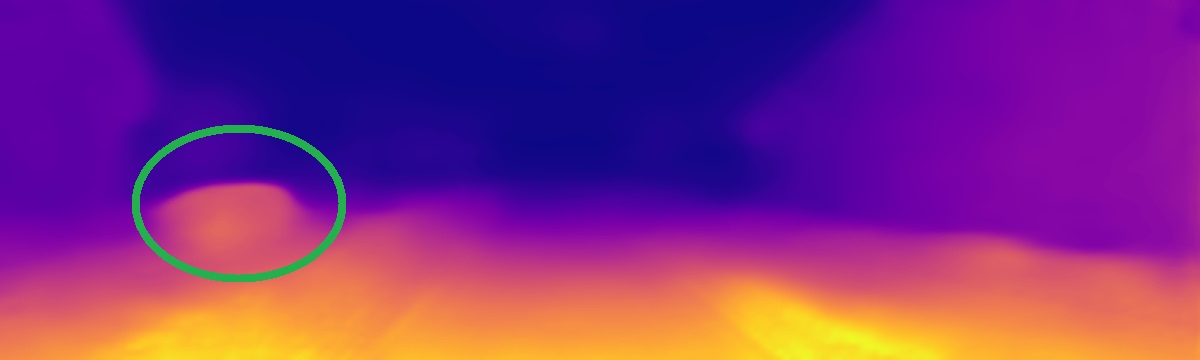} &
\includegraphics[height=\turnheightnew]{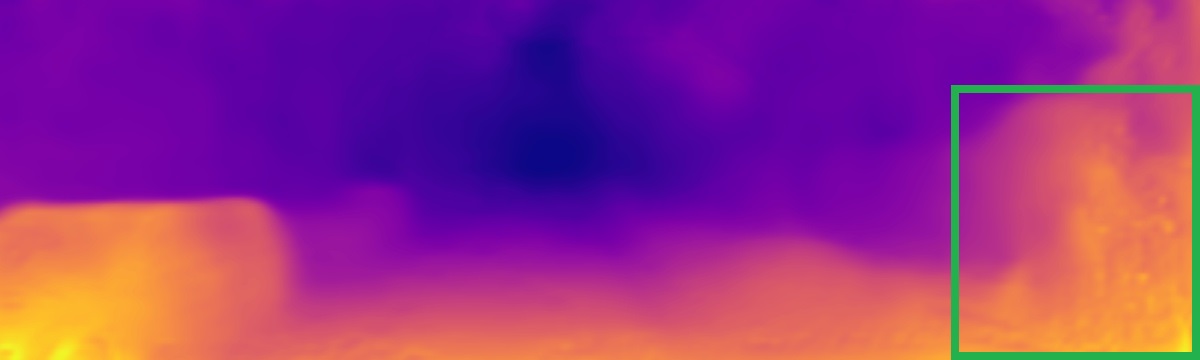} &
\includegraphics[height=\turnheightnew]{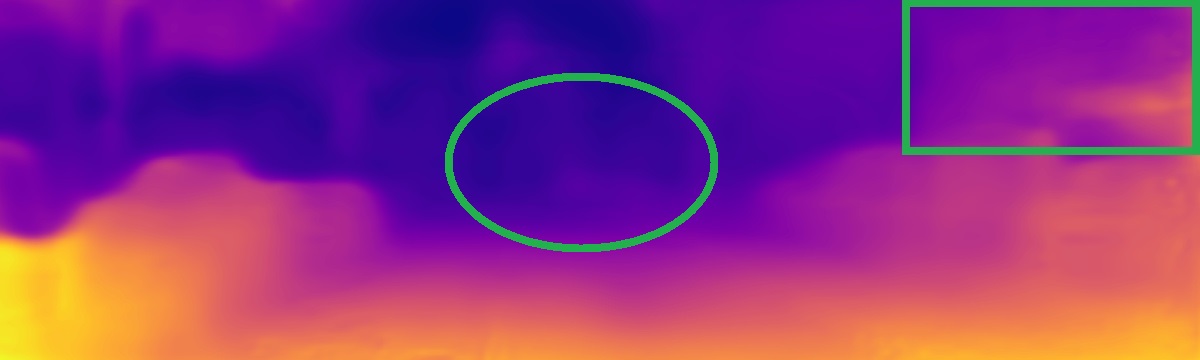} &
\includegraphics[height=\turnheightnew]{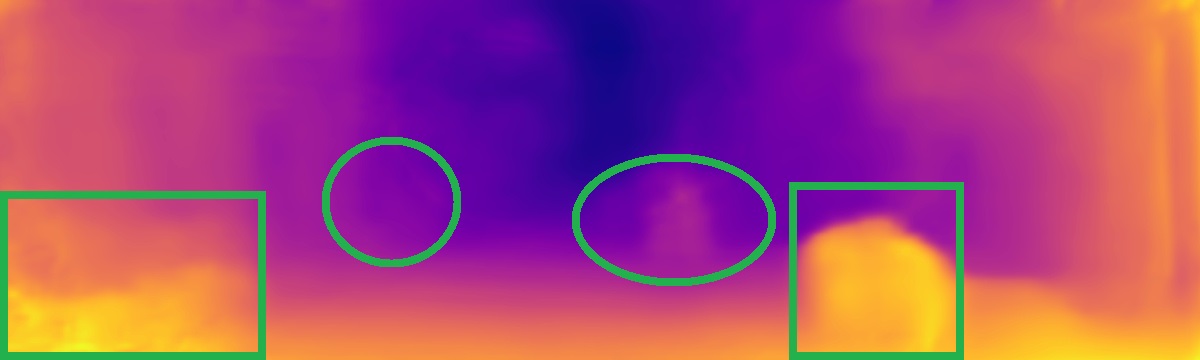}\\

{\rotatebox{90}{\hspace{1.5mm}\scriptsize
{ EPC++\cite{luo2019every}}}} &
\includegraphics[height=\turnheightnew]{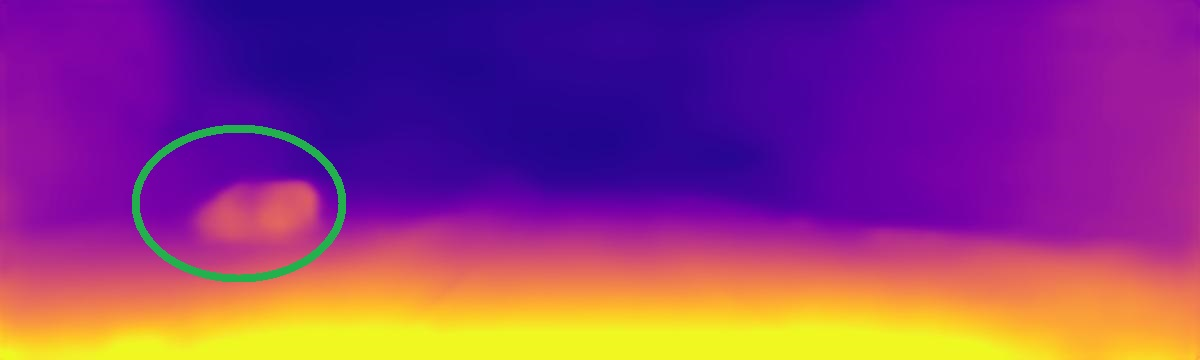} &
\includegraphics[height=\turnheightnew]{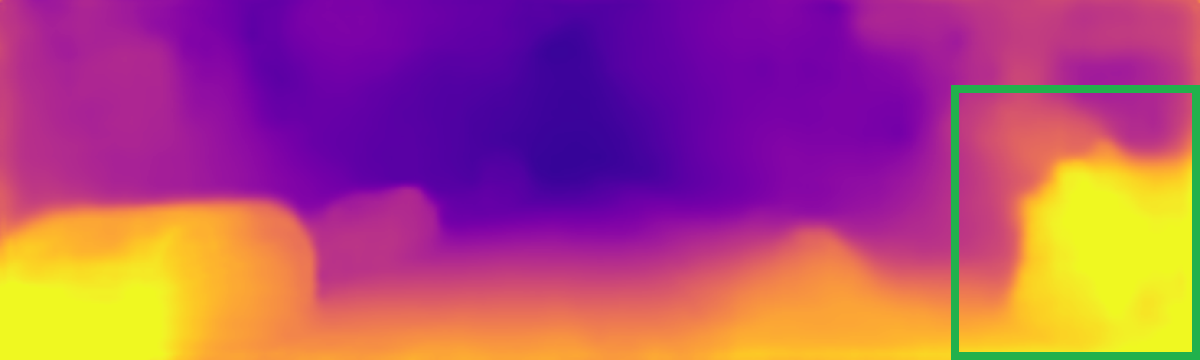} &
\includegraphics[height=\turnheightnew]{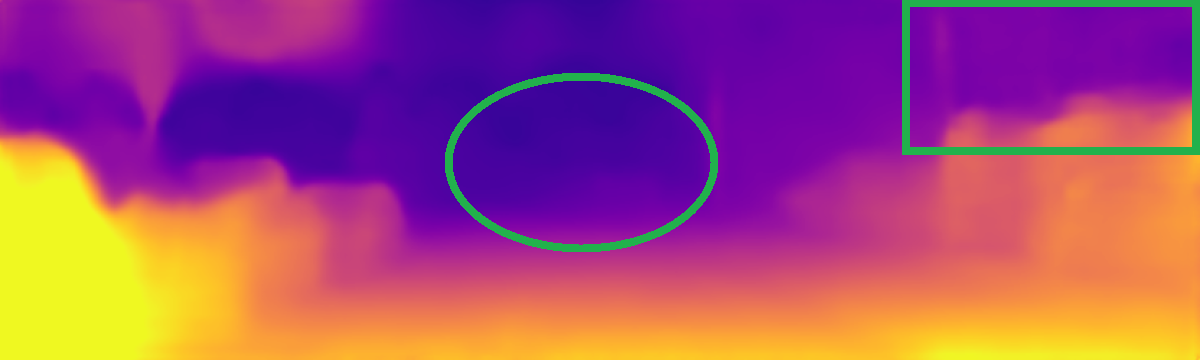} &
\includegraphics[height=\turnheightnew]{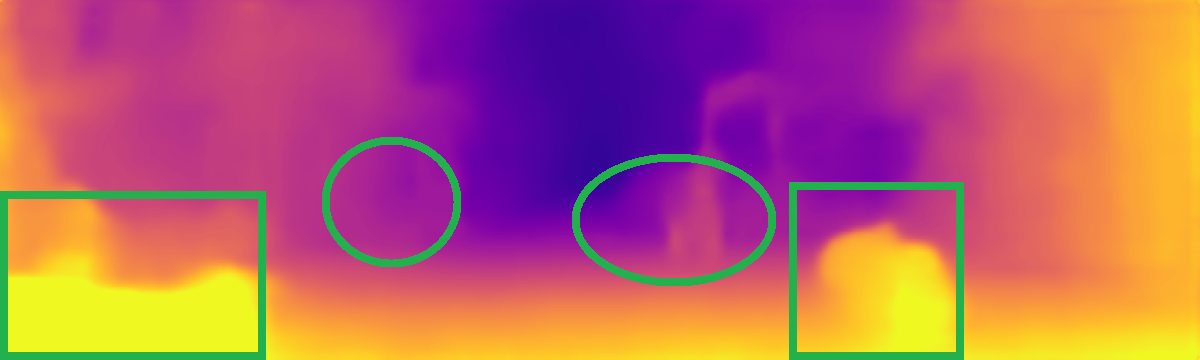}\\

{\rotatebox{90}{\hspace{0mm}\scriptsize
{Struct2depth\cite{casser2019depth}}}} &
\includegraphics[height=\turnheightnew]{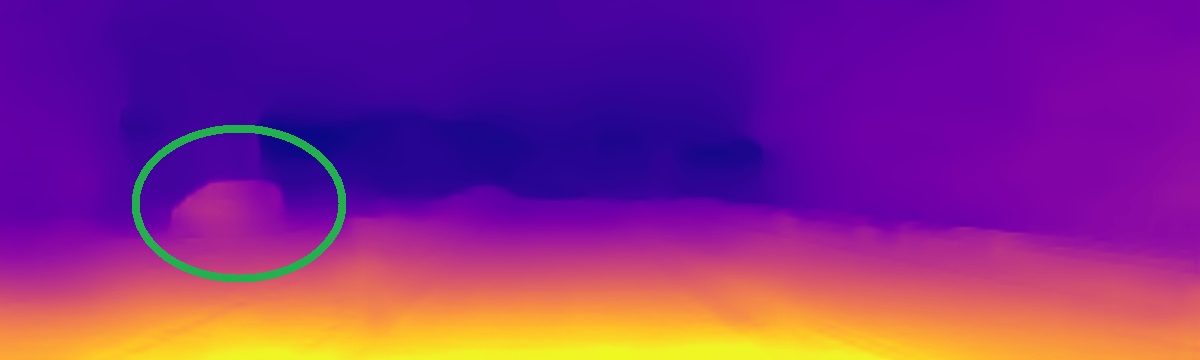} &
\includegraphics[height=\turnheightnew]{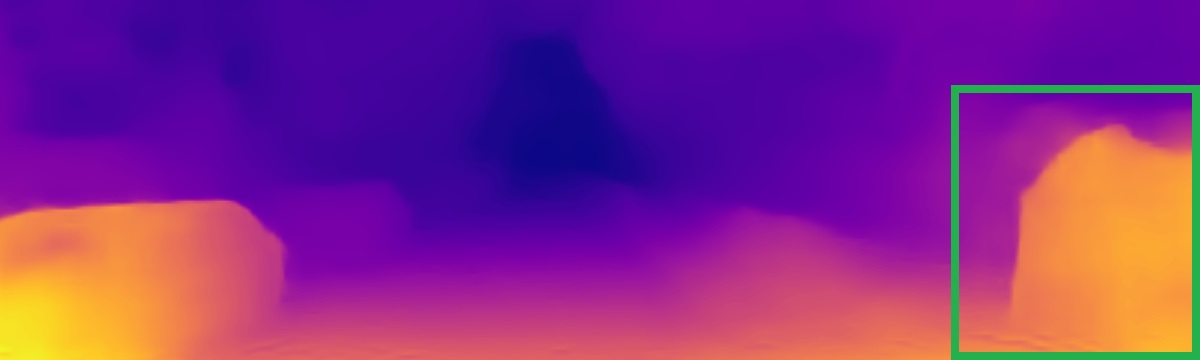} &
\includegraphics[height=\turnheightnew]{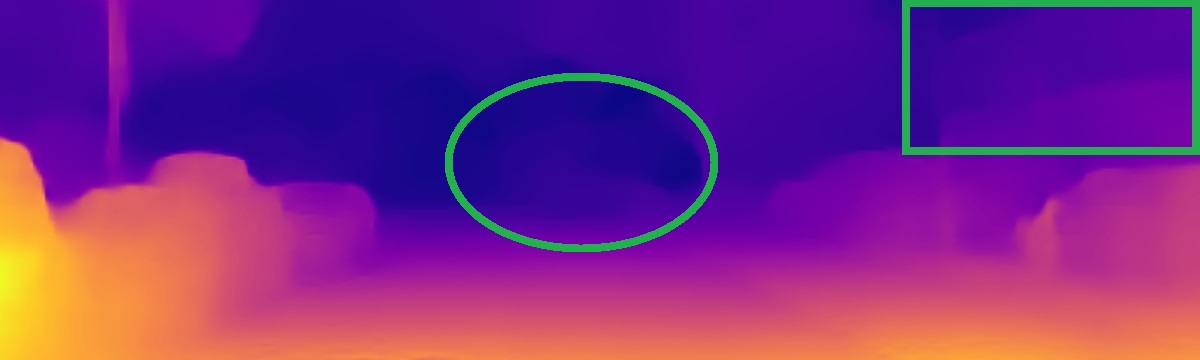} &
\includegraphics[height=\turnheightnew]{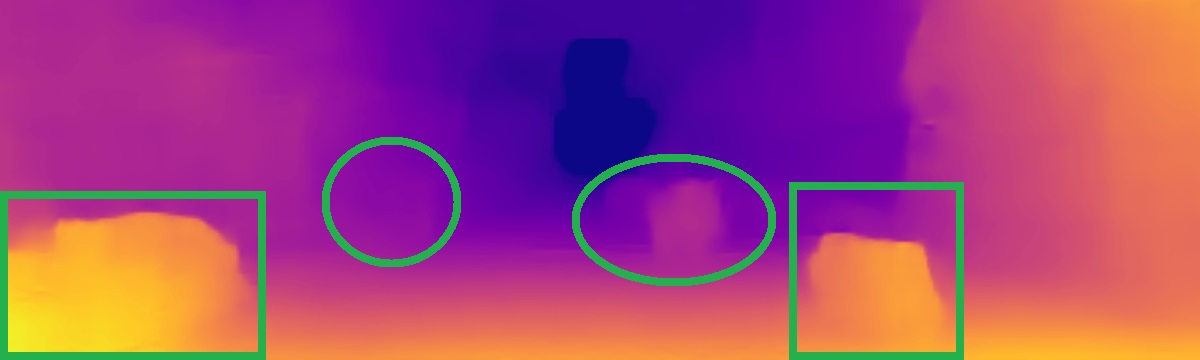}\\

{\rotatebox{90}{\hspace{0mm}\scriptsize
{DOPlearning~\cite{wang2020unsupervised}}}} &
\includegraphics[height=\turnheightnew]{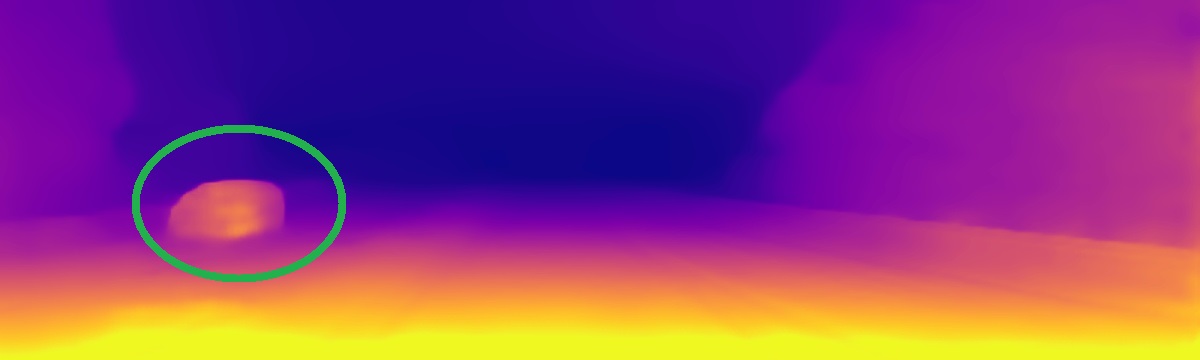} &
\includegraphics[height=\turnheightnew]{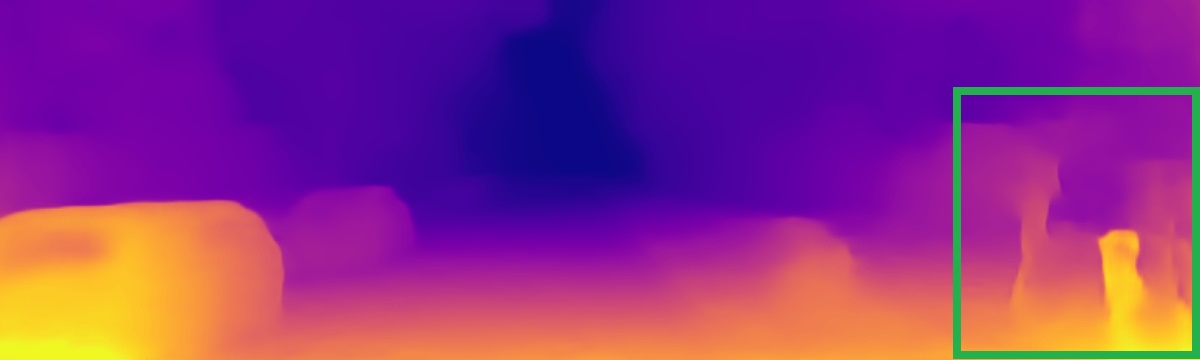} &
\includegraphics[height=\turnheightnew]{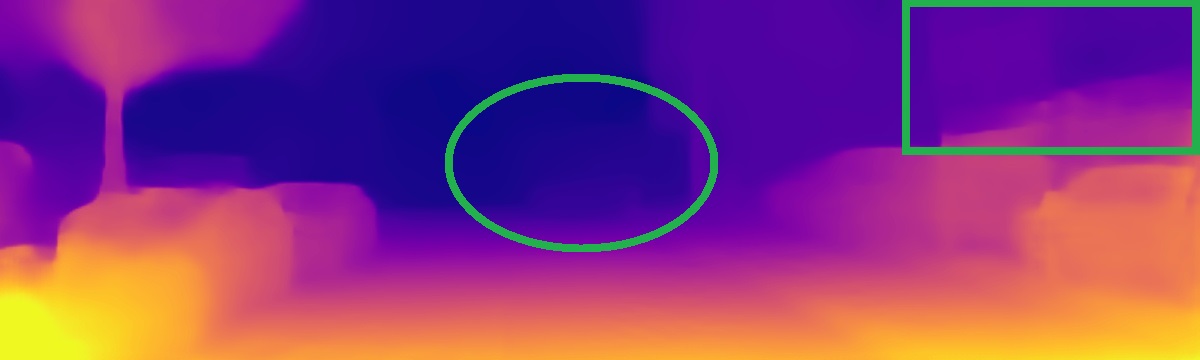} &
\includegraphics[height=\turnheightnew]{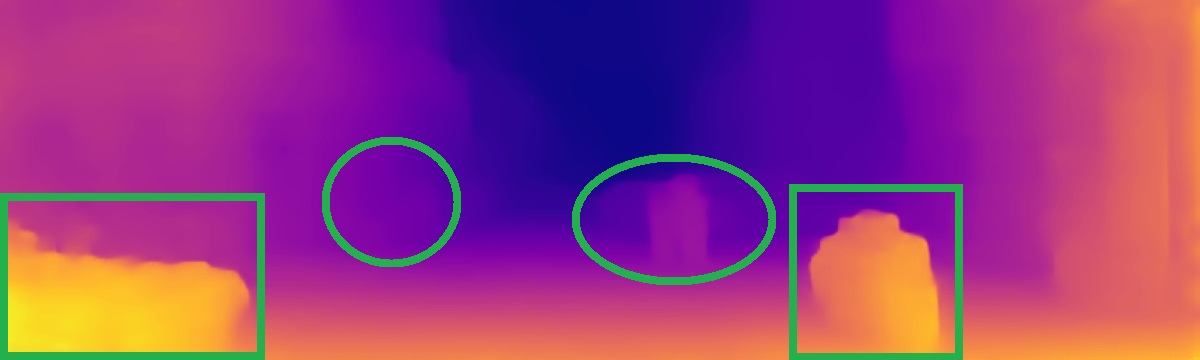}\\

{\rotatebox{90}{\hspace{-0.2mm}\scriptsize
{Monodepth2\cite{godard2019digging}}}} &
\includegraphics[height=\turnheightnew]{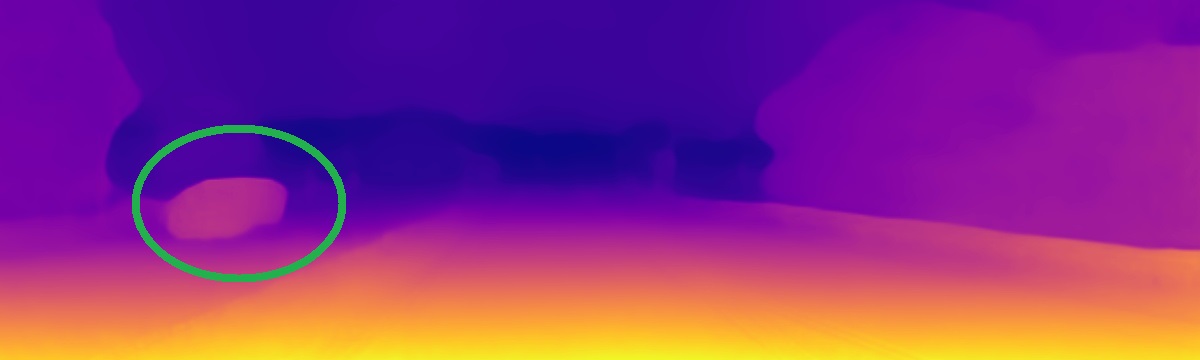} &
\includegraphics[height=\turnheightnew]{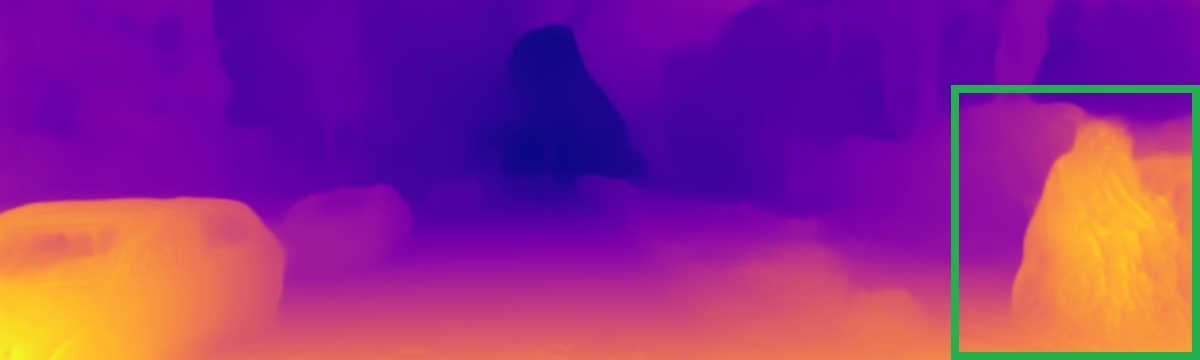} &
\includegraphics[height=\turnheightnew]{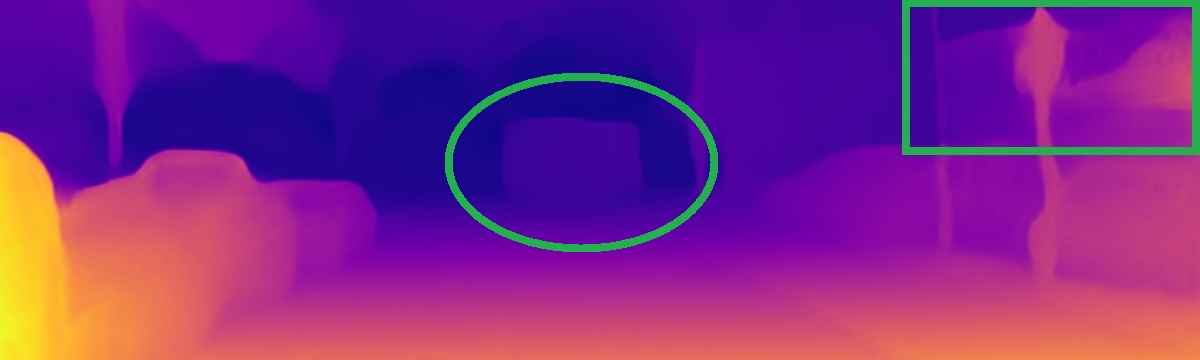} &
\includegraphics[height=\turnheightnew]{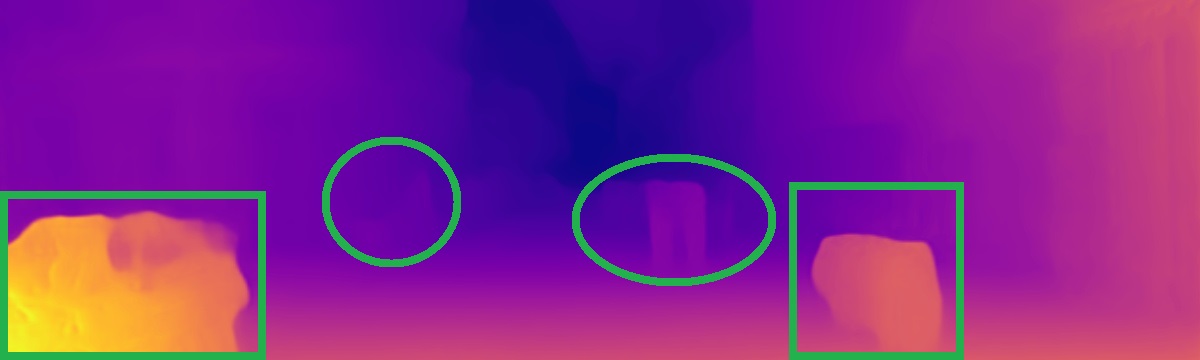}\\

{\rotatebox{90}{\hspace{-0.2mm}\scriptsize
{PackNet-SfM\cite{guizilini20203d}}}} &
\includegraphics[height=\turnheightnew]{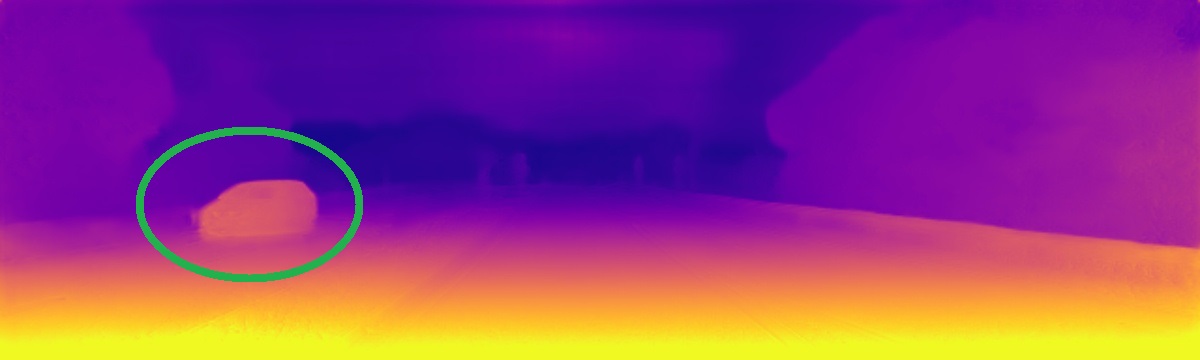} &
\includegraphics[height=\turnheightnew]{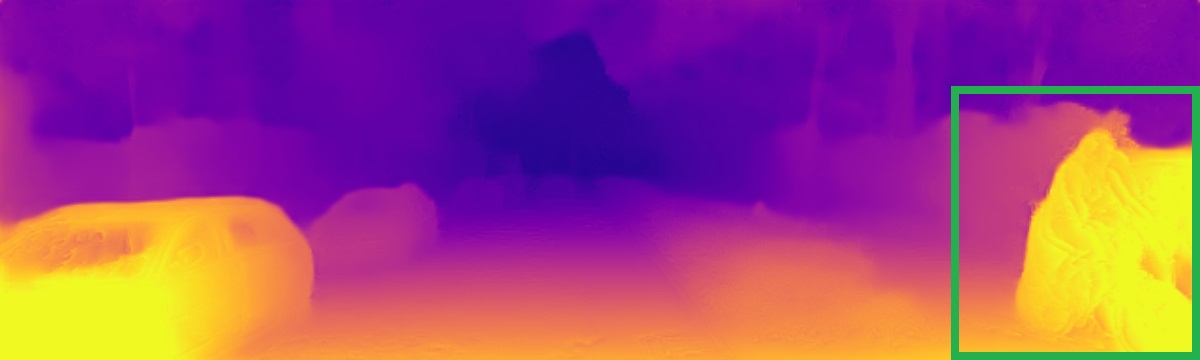} &
\includegraphics[height=\turnheightnew]{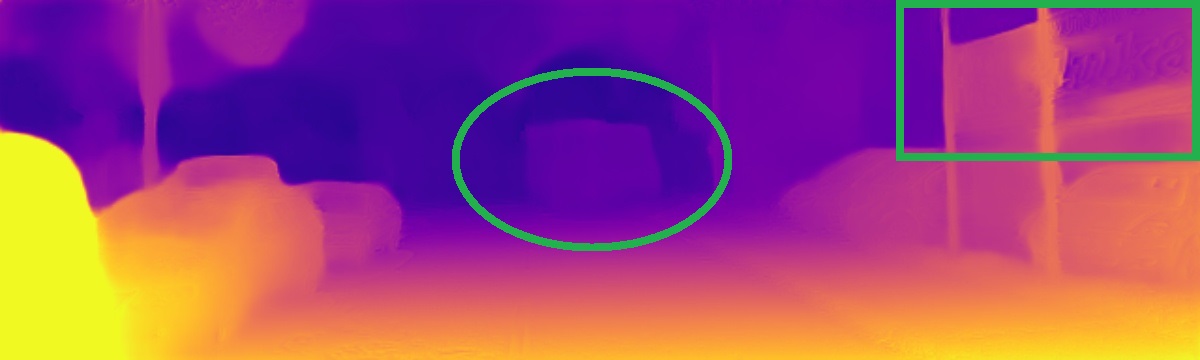} &
\includegraphics[height=\turnheightnew]{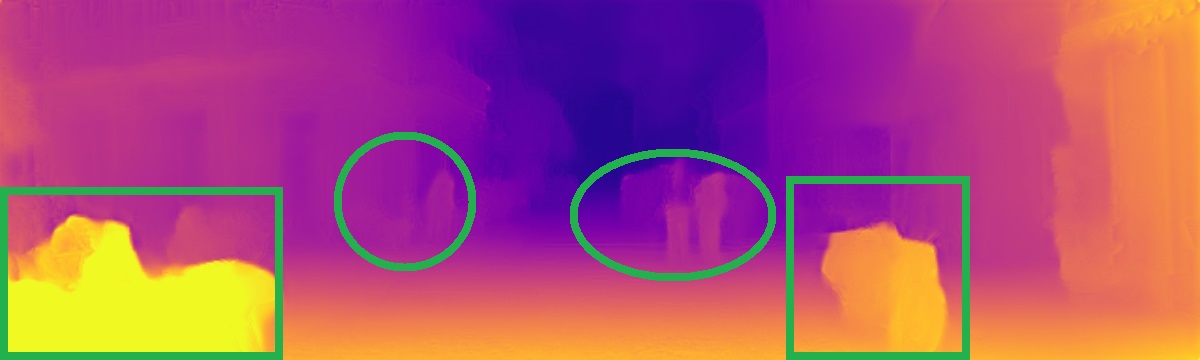}\\

{\rotatebox{90}{\hspace{1.2mm}\scriptsize
{DiPE (Ours)}}} &
\includegraphics[height=\turnheightnew]{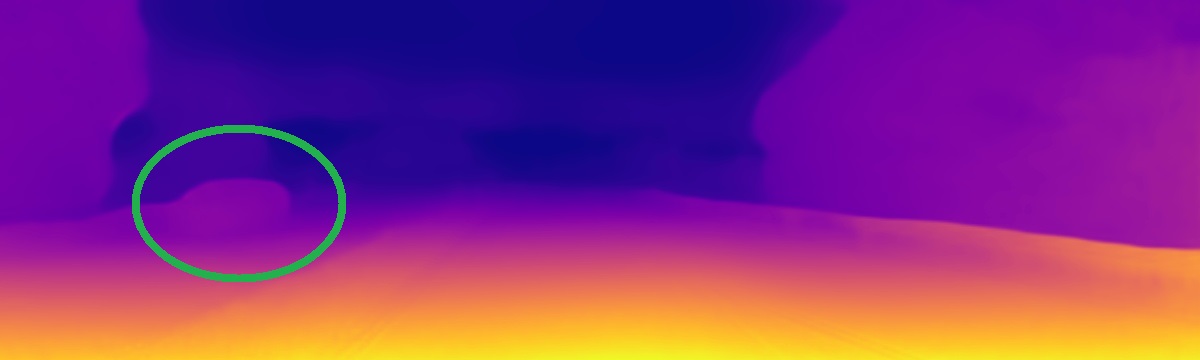} &
\includegraphics[height=\turnheightnew]{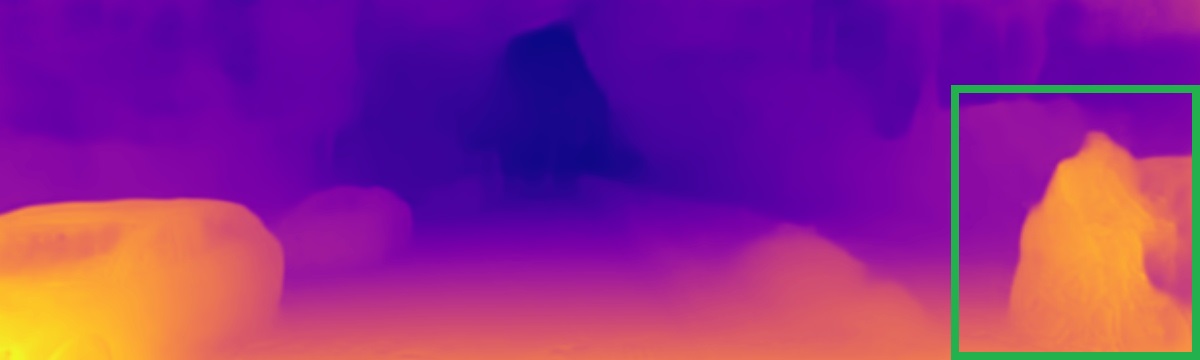} &
\includegraphics[height=\turnheightnew]{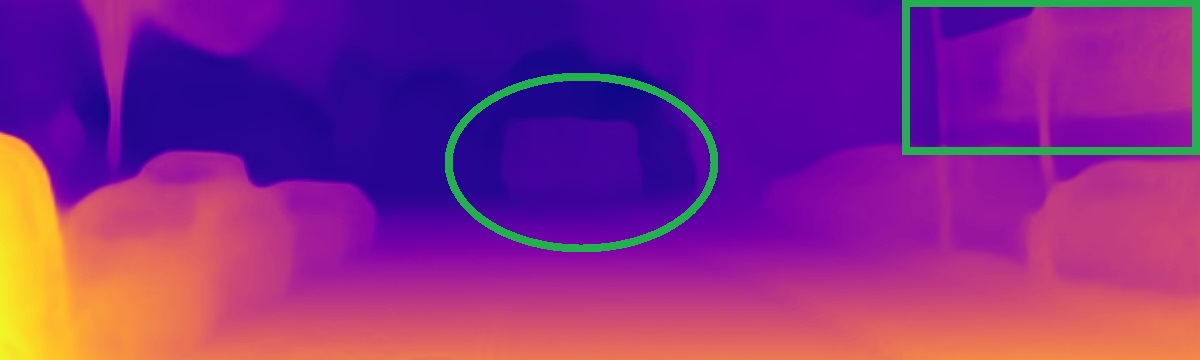} &
\includegraphics[height=\turnheightnew]{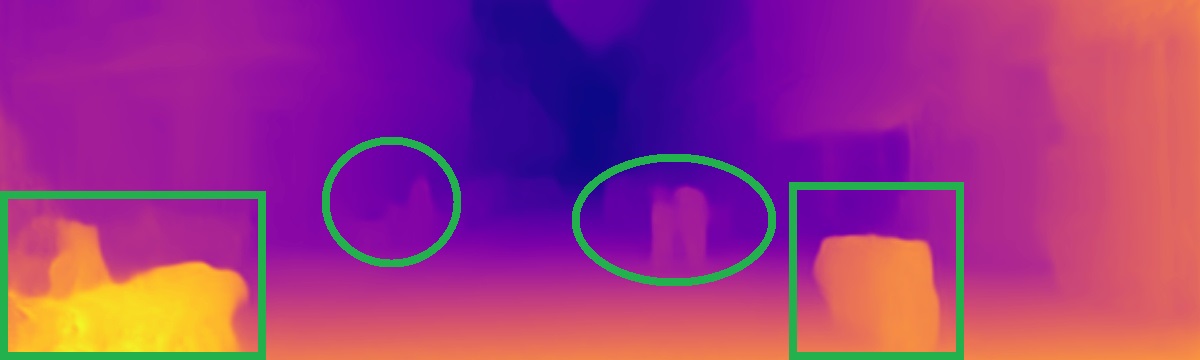}\\

\end{tabular}
 }
\caption{\textbf{Qualitative comparison.} Our model DiPE produces very high-quality depth maps, and it reduces many artifacts due to occlusion and scene dynamics. More importantly, recent state-of-the-art models including joint learning with optical flow \cite{yin2018geonet, luo2019every}, modeling object motion \cite{casser2019depth}, hand ling occlusion with masks and joint learning with optical flow simultaneously~\cite{wang2020unsupervised} and auto-masking moving objects \cite{godard2019digging}, underestimate the depth for the objects moving in the opposite direction, \eg oncoming cars, while our DiPE succeeds (the first column). Zoom in for a better view.}
\label{fig:comparison}
\vspace{-12pt}
\end{figure*}

\begin{table*}[htp]
  \centering
  \caption{\textbf{Quantitative Results.} All methods are trained and evaluated on the Eigen split \cite{eigen2014depth} of the KITTI dataset \cite{geiger2013vision}. }
\label{tab:kitti_eigen}
\vspace{-3pt}
\resizebox{0.88\textwidth}{!}{
\begin{threeparttable}

  \begin{tabular}{l|c|c|cccc|ccc}
  \toprule
  \multirow{2}{*}{Method}  & \multirow{2}{*}{Train} & \multirow{2}{*}{Resolution} & \multicolumn{4}{c|}{Error metric $\downarrow	$} & \multicolumn{3}{c}{Accuracy metric $\uparrow	$}\\
  \cline{4-7}
  \cline{8-10}
   &  & & Abs Rel & Sq Rel & RMSE  & RMSE log & $\delta < 1.25 $ & $\delta < 1.25^{2}$ & $\delta < 1.25^{3}$  \\
  \hline
  \hline
Eigen~{\etal}~\cite{eigen2014depth} & D & 576$\times$271 & 0.203 & 1.548 & 6.307 & 0.282 & 0.702 & 0.890 & 0.890\\
Liu~{\etal}~\cite{liu2015learning} & D & 640$\times$192 & 0.201 & 1.584 & 6.471 & 0.273 & 0.680 & 0.898 & 0.967\\
Kuznietsov~{\etal}~\cite{kuznietsov2017semi} & DS & 621$\times$187 & 0.113 & 0.741 & 4.621 & 0.189 & 0.862 & {0.960} & {0.986}\\
DORN~\cite{fu2018deep} & D & 1241$\times$385 & \textbf{0.072}&  \textbf{0.307} & \textbf{2.727} & \textbf{0.120} & \textbf{0.932} & \textbf{0.984} & \textbf{0.994}\\ 
\hline
\hline
Garg~{\etal}~\cite{garg2016unsupervised} & S & 620$\times$187 &  0.152 & 1.226 & 5.849 & 0.246 & 0.784 & 0.921 & 0.967\\
Monodepth R50~\cite{godard2017unsupervised}\textdagger & S & 512$\times$256 & 0.133 & 1.142 & 5.533 & 0.230 & 0.830 & 0.936 & 0.970\\
SuperDepth~\cite{pillai2019superdepth} & S & 1024$\times$384 & 0.112 & 0.875 & {4.958} & {0.207} & 0.852 & 0.947 & {0.977}\\
Monodepth2~\cite{godard2019digging}  & S & 640$\times$192 & 
{0.109} & {0.873} &  {4.960} &   {0.209} &   {0.864} &   {0.948} &  {0.975} \\
Tian~{\etal}~\cite{tian2021depth} & S & 1216$\times$352 & \textbf{0.095} & \textbf{0.601} &  \textbf{4.128} &   \textbf{0.176} &   \textbf{0.908} &   \textbf{0.976} &  \textbf{0.991} \\
\hline
\hline
SfMLearner~\cite{zhou2017unsupervised}\textdagger & M & 416$\times$128 & 0.183 & 1.595 & 6.709 & 0.270 & 0.734 & 0.902 & 0.959\\
Vid2Depth~\cite{mahjourian2018unsupervised} & M & 416$\times$128 & 0.163 & 1.240 & 6.220 & 0.250 & 0.762 & 0.916 & 0.968\\
DF-Net~\cite{zou2018df} & M & 576$\times$160 & 0.150 & 1.124 & 5.507 & 0.223 & 0.806 & 0.933 & 0.973\\
GeoNet~\cite{yin2018geonet}\textdagger & M & 416$\times$128 & 0.149 & 1.060 & 5.567 & 0.226 & 0.796 & 0.935 & 0.975\\
DDVO~\cite{wang2018learning} & M & 416$\times$128 & 0.151 & 1.257 & 5.583 & 0.228 & 0.810 & 0.936 & 0.974\\
EPC++~\cite{luo2019every} & M & 832$\times$256 & 0.141 & 1.029 & 5.350 & 0.216 & 0.816 & 0.941 & 0.976\\
Struct2depth `(M)'~\cite{casser2019depth} & M & 416$\times$128 & 0.141 & {1.026} & 5.291 &  0.215 & 0.816 & 0.945 & {0.979}\\
DOPlearning~\cite{wang2020unsupervised} & M & 832$\times$256 & 0.140 & {1.068} & 5.255 &  0.217 & 0.827 & 0.943 & {0.977}\\
SC-SfMLearner~\cite{bian2019depth} & M & 832$\times$256  & 0.137 & 1.089 & 5.439 & 0.217 & 0.830 & 0.942 & 0.975\\
Gordon~\etal \cite{Gordon_2019_ICCV} & M & 416$\times$128 & 0.128 & 0959 & 5.230 &  0.212 & 0.845 & 0.947 & {0.976}\\
Monodepth2~\cite{godard2019digging} & M & 640$\times$192& {0.115} & {0.903} & {4.863} &  {0.193} &   {0.877} & {0.959} &   {0.981} \\ 
PackNet-SfM~\cite{guizilini20203d} & M & 640$\times$192&
 {\bf 0.111} &   {\bf 0.785} &   {\bf 4.601} &   {\bf 0.189} &   {0.878} &   {\bf 0.960} &   {\bf 0.982} \\ 
\textbf{DiPE} (Ours) & M & 640$\times$192 &   
 {0.112} & {0.875} & {4.795} & {0.190} & {\bf 0.880} & {\bf 0.960} & {0.981} \\ 
  \bottomrule
  \end{tabular}

\begin{tablenotes}[flushleft]
\footnotesize   
\item Note: Three categories of methods that perform training with the depth, stereo images, and monocular video frames are compared. In each category, the best results are in \textbf{bold}. \textbf{Legend:} D -- depth supervision; S -- unsupervised stereo supervision; M -- unsupervised monocular supervision; \textdagger -- newer results from the respective  implementations.
\end{tablenotes}
\end{threeparttable}
}
\end{table*}

Besides, an edge-aware smoothness loss is usually applied in unsupervised training.  We use the one from \cite{godard2019digging},
\begin{equation}
L_{es} = mean\left( \left | \partial_x d^*_t   \right | e^{-\left | \partial_x I_t \right |} + \left | \partial_y d^*_t   \right | e^{-\left | \partial_y I_t \right |} \right), 
\label{eq:smooth}
\end{equation} 
where $d^* = d / \overline{d}$ is the mean-normalized inverse depth from \cite{wang2018learning} to discourage shrinking of the estimated depth. {Both losses are applied in $4$ scales to avoid gradient locality. }

\section{Experiments}
\label{sc:exp}

This section shows experiments to validate the effectiveness of our approach and present some critical observations and discussions from these experiments.

\subsection{Implementation Details}
\label{sc:impl}

We implement the proposed approach based on Monodepth2 \cite{godard2019digging} and maintain the most experimental settings. The depth CNN is a fully convolutional encoder-decoder network with an input/output resolution of $640\times192$. The Pose CNN is a typical CNN with a fully connected layer to regress the 6-Dof relative camera pose. In all of the experiments,  both networks use a ResNet18 \cite{he2016deep} pre-trained on ImageNet \cite{deng2009imagenet} as backbone.
In depth estimation experiments, as Monodetph2 \cite{godard2019digging} does, we only use the $2$ nearby frames ($\mathcal{S} = \{t-1, t+1\}$) and the pair-input Pose (Fig.~\ref{fig:framework}). But in ego-motion estimation expeiments, we also experiment with the all-input Pose CNN with the $4$ nearby frames ($\mathcal{S} = \{t-2, t-1, t+1, t+2\}$) and $2$ frames ($\mathcal{S} = \{t-1, t+1\}$) as \cite{zhou2017unsupervised, mahjourian2018unsupervised}, respectively.

We conventionally set the hyper-parameters $\eta$, $\lambda$, and $e$ in the final loss function as $1$, $0.001$, and $0.5$, respectively.  By examining several values ($1.0$, $0.5$, $0.25$ and $0.125$) using the the validation set, we set factor $f$ of the weighted multi-scale scheme  as $0.25$. We train the models for 20 epochs using Adam \cite{kingma2014adam}. As our weighted multi-scale scheme consumes less memory, the models are trained with a bigger batch size of 16 than 12 in Monodepth2, and the training takes only 9 hours on a single Titan Xp while Monodepth2 uses 12 hours. DiPE uses an initial learning rate of  $10^{-4}$ but divides it by 5 after 15 and 18 epochs. As the outlier masking further reduces the pixels for training, decreasing the learning rate can help DiPE converge better. 
Monodepth2 uses the same intrinsic parameters for all training samples by approximating the camera's principal point to the image center and averaging the focal length on the whole dataset. To be more accurate, we use the calibrated intrinsic parameters for every training sample. When performing horizontal flips in data augmentation, {the horizontal coordinate of the principal point $f_x$ changes into $w - f_x$, where $w$ is the width of the image.}

\subsection{KITTI Eigen Split}
\label{sc:eigen}

The Eigen split \cite{eigen2014depth} of the KITTI dataset \cite{geiger2013vision} is commonly used in monocular depth estimation experiments. Following Zhou~\etal~\cite{zhou2017unsupervised}, we use a subset of the training set that contains no static frames. There are 39,810, 4,424, and 697 samples for training, validation, and test. 
To better monitor the training process, we evaluate about one-tenth (432) of the validation set for every epoch rather than evaluate a batch of validation samples for specific steps (200 or 2000 in Monodepth2~\cite{godard2019digging}). 
In evaluation, every predicted depth map is aligned to the ground truth depth map by multiplying the median value ratio \cite{zhou2017unsupervised} as other unsupervised methods. We also adopt the conventional metrics and cropping region in \cite{eigen2014depth}, and the standard depth cap $80m$ \cite{godard2017unsupervised}.

We quantitatively and qualitatively compare the results of our model and other state-of-the-art methods. 
The quantitative results are shown in Tab.~\ref{tab:kitti_eigen} and the results of other methods are taken from the corresponding papers. 
The comparison is mainly among the unsupervised monocular training methods, but some supervised and unsupervised stereo training models are also included. 
DiPE archives the state-of-the-art performance, as it markedly outperforms the recent Monodepth2 \cite{godard2019digging} {and performs approximately to the more recent PackNet-SfM~\cite{guizilini20203d} which adopts a much more complex Depth CNN}. Also, DiPE has comparable or even better performance to the models of the other two categories. 
Fig~\ref{fig:comparison} shows the qualitative comparison among the predicted depth maps by DiPE and many state-of-the-art unsupervised monocular training methods. The predicted depth maps of other models are either shared by the authors or obtained by running the authors' codes. DiPE handles the scene dynamics and artifacts better. %The attached video provides more results on the oncoming vehicles.

\begin{table}[!ht]
  \centering
\caption{\textbf{KITTI single depth prediction benchmark~\cite{kittidepthserver}. }}
\label{tab:kitti_eval_server}

\resizebox{0.98\columnwidth}{!}{
\begin{threeparttable}
  \begin{tabular}{l|c|cccc}
  \toprule
  \multirow{2}{*}{Method} & \multirow{2}{*}{Train} & \multicolumn{4}{c}{Error metric $\downarrow	$} \\
  \cline{3-6}
   &   & SILog & sqErrorRel & absErrorRel  & iRMSE log  \\
  \hline
  \hline
DHGRL~\cite{zhang2018deep}& D & 15.47 & 4.04  & 12.52  & 15.72 \\ 
HBC~\cite{jiang2019hierarchical} & D & 15.18 & 3.79  & 12.33  & 17.86 \\
CSWS~\cite{li2018monocular}& D & 14.85 & 3.48  & 11.84  & 16.38 \\
APMoE~\cite{kong2018pixel} & D & 14.74 & 3.88  & 11.74  & 15.63 \\
DABC~\cite{li2018deep}& D & 14.49 & 4.08  & 12.72  & 15.53 \\
DORN~\cite{fu2018deep} & D & {\bf 11.77} & {\bf 2.23}  & {\bf 8.78}  & {\bf 12.98} \\
\hline
\hline
Monodepth~\cite{godard2017unsupervised} & S & 22.02	& 20.58 & 	17.79 &	21.84\\
LSIM \cite{goldman2019lsim}& S & 17.92	& 6.88 & 	14.04 & 17.62\\
Monodepth2~\cite{godard2019digging}& M & 15.57 & 4.52 & 12.98 & 16.70 \\
{\bf DiPE (Ours)} & M & {\bf 14.84} & {\bf 4.04} & {\bf 12.28} & {\bf 15.69} \\
\bottomrule
\end{tabular}

\begin{tablenotes}[flushleft]
\footnotesize              
\item Note: Comparison of our DiPE model to some fully supervised and stereo image based unsupervised methods, as well as Monodepth2 \cite{godard2019digging} on KITTI single depth prediction benchmark \cite{kittidepthserver}. D represents models trained with ground truth depth supervision, whereas M and S are monocular and stereo supervision, respectively.
\end{tablenotes}
\end{threeparttable}

}
\vspace{-5pt}
\end{table}

\subsection{KITTI Single Image Depth Benchmark}

\begin{table*}[t!]
  \centering
  \caption{{\bf Ablation Study.} Results for two groups of models with monocular training on the Eigen split \cite{eigen2014depth} of the KITTI dataset~\cite{geiger2013vision}.}
  \label{tab:kitti_eigen_ablation}
\resizebox{0.98\textwidth}{!}{
\begin{threeparttable}

\footnotesize
 \begin{tabular}{l|l||c|c|c||c|c|c|c|c|c|c}
 \toprule
 & & \multirow{2}{*}{\begin{tabular}{@{}c@{}}Full res. \\ multi-scale\end{tabular}}
 & \multirow{2}{*}{\begin{tabular}{@{}c@{}}Weighted \\ multi-scale\end{tabular}} 
 & \multirow{2}{*}{\begin{tabular}{@{}c@{}}Outlier \\ masking\end{tabular}} 
 & \multicolumn{4}{c|}{Error metric $\downarrow	$} & \multicolumn{3}{c}{Accuracy metric $\uparrow	$}\\
  \cline{6-9}
  \cline{10-12}
   &&&&   & Abs Rel & Sq Rel & RMSE  & RMSE log & $\delta < 1.25 $ & $\delta < 1.25^{2}$ & $\delta < 1.25^{3}$   \\
  \hline
  \hline
  (a) & Baseline I & & & &  0.141  &   1.559  &   5.504  &   0.227  &   0.847  &   0.945  &   0.973  \\
  \clinegrayzero
  & w/o  Wt. m.s. &  & & \checkmark &    0.133  &   1.549  &   5.420  &   0.215  &   0.859  &   0.951  &   0.977  \\
  \clinegrayone
  & w/o Ol. m. & & \checkmark &  &   0.132  &   1.413  &   5.372  &   0.215  &   0.856  &   0.950  &   0.976  \\
  \clinegrayone
  & DiPE  & & \checkmark & \checkmark & {\bf 0.123} & {\bf 1.206} & {\bf 5.163} & {\bf 0.202}  & {\bf 0.870} & {\bf 0.955}  &   {\bf 0.978}  \\
 \hline
 \hline
(b) & Baseline II & & &  &  0.120  &   0.927  &   4.938  &   0.198  &   0.868  &   0.956  &   0.980  \\
\clinegrayone
  & + F.r. m.s. & \checkmark &  &  &   0.116  &   0.939  &   4.905  &   0.194  &   0.874  &   0.957  &    {\bf 0.981}  \\
  \clinegrayone
  & + F.r. m.s. + Ol. m. & \checkmark &  & \checkmark &   0.116  &   0.959  &   4.892  &   0.193  &   0.876  &   0.958  &   {\bf 0.981}  \\
  \clinegrayzero
 & w/o Wt. m.s. & & & \checkmark &   0.117  &   0.921  &   4.915  &   0.196  &   0.871  &   0.957  &   {\bf 0.981}  \\
  \clinegrayone
  & w/o Ol. m. & & \checkmark &  &   0.115  &   0.910  &   4.865  &   0.193  &   0.876  &   0.958  &   0.980  \\
  \clinegrayone
  & DiPE & & \checkmark & \checkmark & {\bf 0.112} &   {\bf 0.875} &   {\bf 4.795} &   {\bf 0.190} &   {\bf 0.880} &   {\bf 0.960} &   {\bf 0.981} \\ 
  \bottomrule
  \end{tabular}

\begin{tablenotes}[flushleft]
\footnotesize              
\item Note: 
      \textbf{(a)} The ablation experiments on the first baseline model, which adopts the conventional multi-scale method and no masking techniques. 
      \textbf{(b)} The ablation experiments on the second baseline model, which also adopts the principled masking, auto-masking, and minimum reprojection techniques than the first baseline model.  \textbf{Abbreviation:} F.r. m.s.--Full resolution multi-scale scheme; Wt. m.s.--Weighted multi-scale scheme; Ol. m.--Outlier masking technique.
\end{tablenotes}
\end{threeparttable}
}
\vspace{-8pt}
\end{table*}

Following Monodepth2 \cite{godard2019digging}, we also experiment on the KITTI official single image depth prediction benchmark, which uses a different split of the data from the Eigen split \cite{eigen2014depth}.
We train a new model on the training data of the benchmark split. In this split, we have evaluated our method by submitting the predicted depth maps for the test set to the online evaluation server. 
Following Monodepth2 \cite{godard2019digging}, we obtain the scale factor for the test set by making an evaluation on 1000 images of the benchmark split with ground truth, and the resulted scale is the median of the 1,000 scale factors. The results are shown in Tab.~\ref{tab:kitti_eval_server}. 
The results of Monodepth \cite{godard2017unsupervised} and Monodepth2 \cite{godard2019digging} are provided by Monodepth2 \cite{godard2019digging}, and the results of other methods can be found in the evaluation server \cite{kittidepthserver}.  The results illustrate that DiPE can outperform some fully-supervised methods, e.g., DHGRL~\cite{zhang2018deep} and HBC~\cite{jiang2019hierarchical}.

\begin{figure}[t]
\centering
\resizebox{0.95\linewidth}{!}{
\newcommand{\shiftleft}[2]{\makebox[-6pt][r]{\makebox[#1][l]{#2}}}
\newcommand{\imlabel}[2]{\includegraphics[width=0.49\columnwidth]{#1}%
\raisebox{28pt}{\shiftleft{56pt}{\makebox[-2pt][r]{\footnotesize #2}} }}

\centering
\renewcommand{\arraystretch}{0.5}
\begin{tabular}{@{\hskip -1.2mm}c@{\hskip 1.5mm}c}
\imlabel{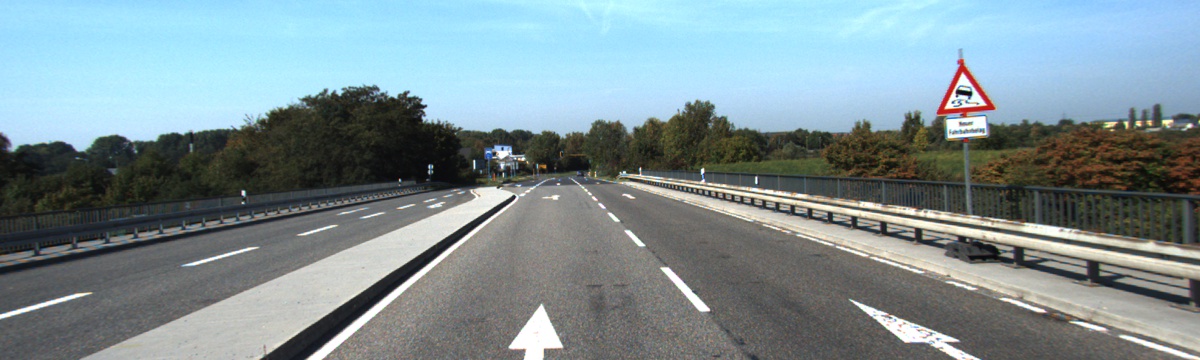}{} &
\imlabel{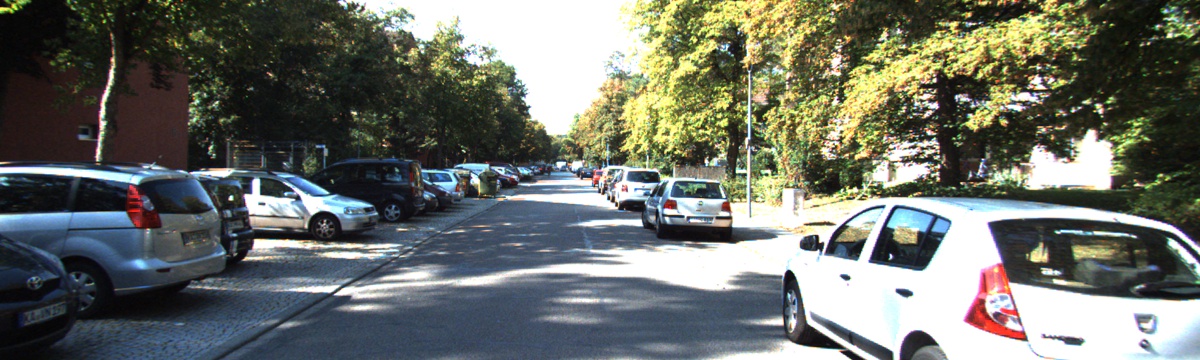}{} \\
\imlabel{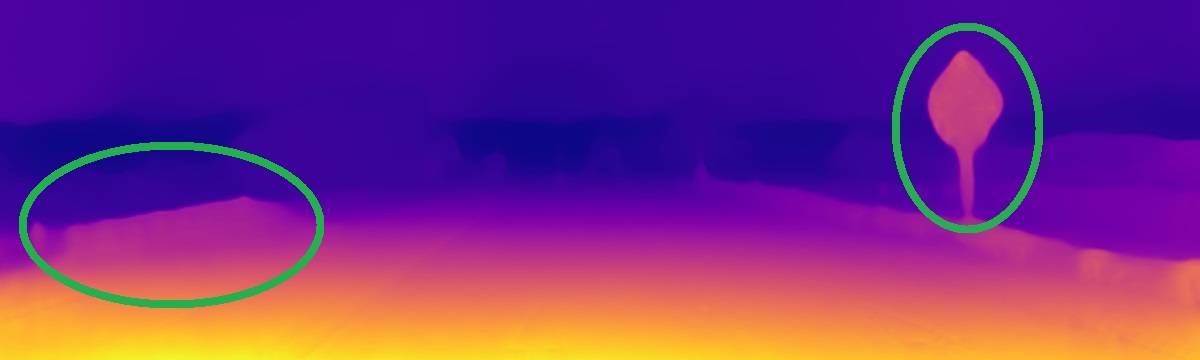}{\textcolor{white}{Monodepth2~\cite{godard2019digging}}} &
\imlabel{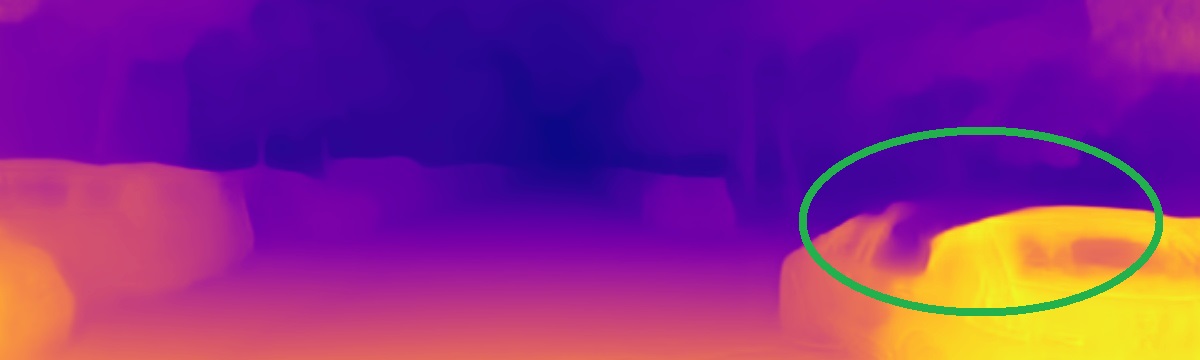}{\textcolor{white}{Monodepth2~\cite{godard2019digging}}} \\
\imlabel{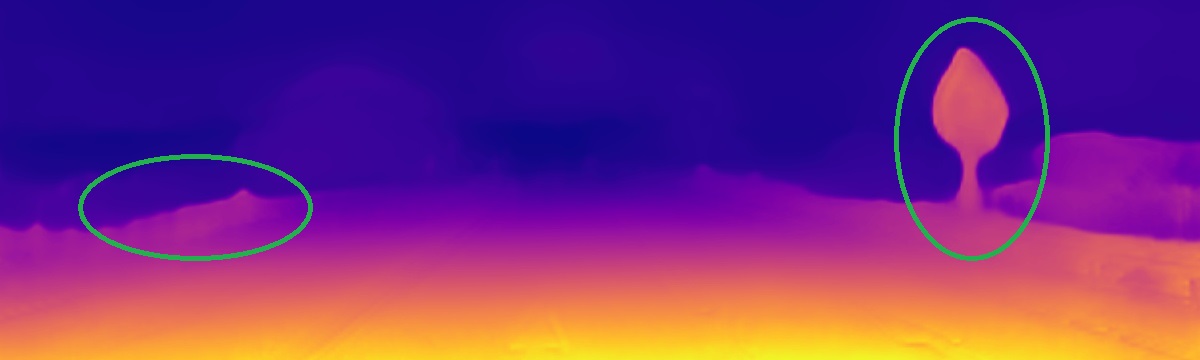}{\textcolor{white}{Baseline II \ \qquad}} &
\imlabel{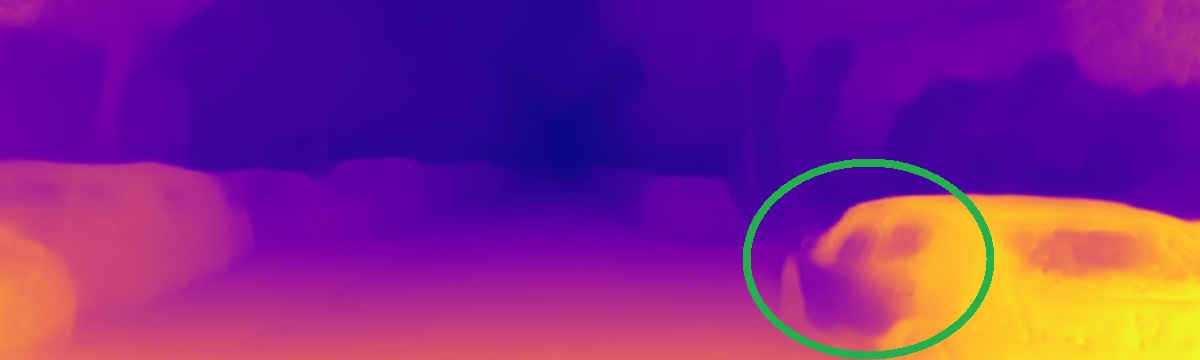}{\textcolor{white}{Baseline II \ \qquad}} \\
\imlabel{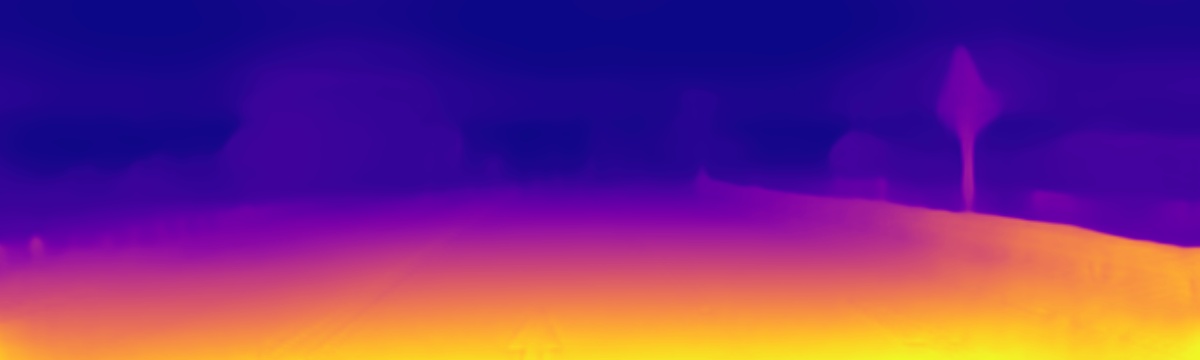}{\textcolor{white}{DiPE  \ \ \qquad \qquad }} &  
\imlabel{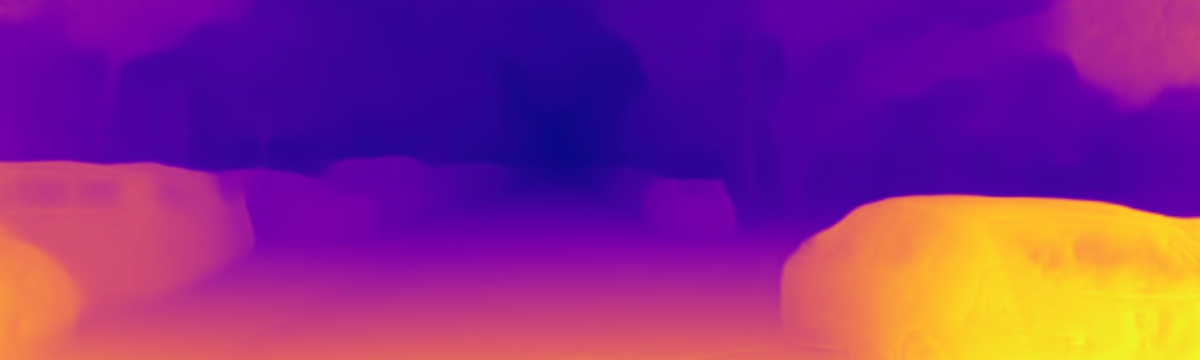}{\textcolor{white}{DiPE  \ \ \qquad \qquad }} \\

\end{tabular}

}
\caption{{\bf Artifacts.} DiPE can solve the artifacts better and succeed in the two failure cases by Monodepth2~\cite{godard2019digging}. }
\label{fig:artifacts}
\vspace{-10pt}
\end{figure}

\subsection{Ablation Study}

We also perform controlled experiments to examine the effectiveness of our contributions. 
We use two baseline models. 
The first model is almost the same as the baseline in Monodepth2 \cite{godard2019digging}, except for the slightly different training scheme. 
To obtain better results, we also adopt a stronger baseline model, which additionally uses the three masking techniques mentioned in Section \ref{sec:objective}, i.e., the principled masking, auto-masking, and minimum reprojection. 
For both baselines, we experiment with four possible combinations of our two contributions, the weighted multi-scale scheme and the outlier masking technique. For the second baseline, we also experiment with the full resolution multi-scale method in Monodepth2 \cite{godard2019digging} to compare with our proposed weighted multi-scale scheme. 
Tab.~\ref{tab:kitti_eigen_ablation} shows the results.

It can be observed that, for both baseline models, our two techniques can obviously improve the performance individually, and the performance gain when they combine together is more than the sum of their separate performance gains, which indicates that the two techniques can collaborate well. Whereas the full resolution multi-scale scheme does not work well with our outlier making technique, as there is almost no additional gain when adding the latter upon the former. Besides, our more efficient weighted multi-scale scheme slightly outperforms the full resolution multi-scale scheme. Furthermore, the weighted multi-scale scheme also helps DiPE address the artifacts better than Monodepth2 \cite{godard2019digging} does. DiPE can handle the two failure cases in Monodepth2, as illustrated in Fig.~\ref{fig:artifacts}.

\subsection{Depth Evaluation on Different Motion Patterns}
\label{sec:dynamic}

\begin{figure}
\centering
\resizebox{\linewidth}{!}{
\newcommand{\shiftleft}[2]{\makebox[-6pt][r]{\makebox[#1][l]{#2}}}
\newcommand{\imlabel}[2]{\includegraphics[width=0.5\columnwidth]{#1}%
\raisebox{28pt}{\shiftleft{82pt}{\makebox[-2pt][r]{\footnotesize #2}} }}

\centering
\renewcommand{\arraystretch}{0.5}
\begin{tabular}{@{\hskip -2mm}c}
\imlabel{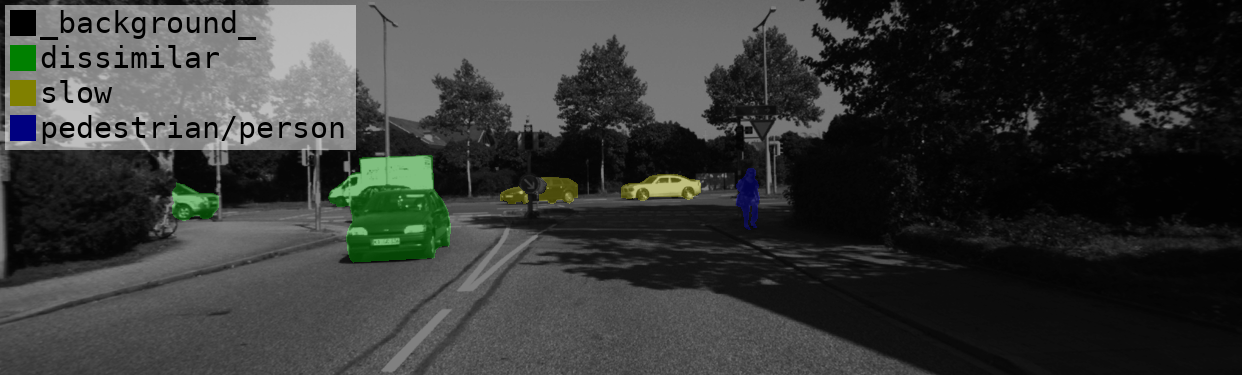}{} \\
\imlabel{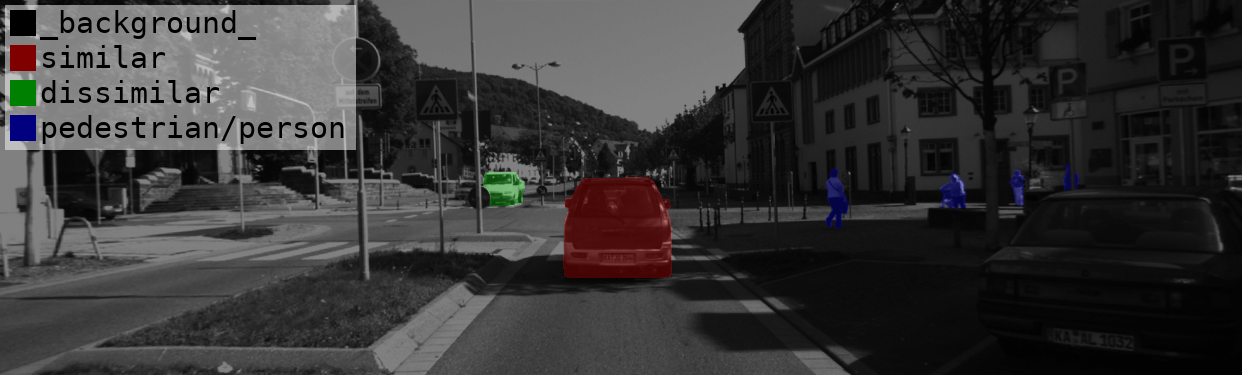}{} \\
\imlabel{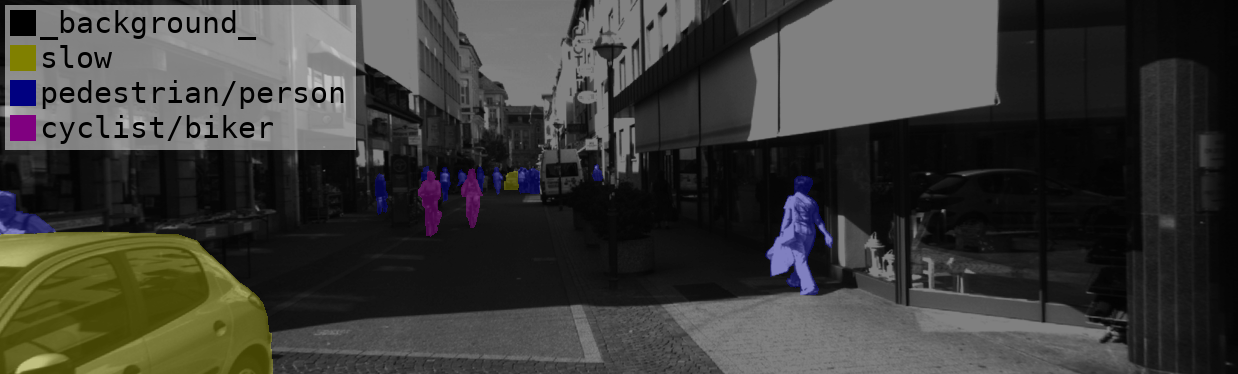}{} \\
%\imlabel{figs/label_sp/693.png}{} \\
\end{tabular}

}
\centering
\caption{{Samples of Labeled Dynamic Objects of Different Patterns.}}
\label{fig:dyn-lbl}
\vspace{-10pt}
\end{figure}

Unlike conventional depth evaluation that counts all the pixels with ground truth, we, for the first time, show evaluation on regions with different motion patterns separately. To perform such evaluation, we, in a pixel-wise manner, labeled the test images of the KITTI Eigen split with different motion patterns with \textit{labelme}~\cite{labelme2016}. We divide the moving objects into five categories, i.e., the similarly moving vehicles, the dissimilarly moving vehicles, the slowly moving vehicles, pedestrians/persons, and cyclists/bikers. Fig.~\ref{fig:dyn-lbl} demonstrates three samples of the labeled dynamic objects. The similarly moving vehicles denote the ones that move at at a speed close to the data-collecting car on the same lane. Such objects are usually predicted to be much further than reality and have been identified and tackled~\cite{luo2019every, casser2019depth, godard2019digging}. 
In contrast, dissimilarly moving vehicles indicate the cars with a significantly different moving pattern from the camera, thus producing bigger photometric errors in between-frame reconstruction. Such vehicles include those moving towards the data-collecting car and moving across in front of the data-collecting car. 
We also distinguish the vehicles at a very slow speed, such as the cars in a traffic jam or moving in a crowded street. Besides vehicles, we also labeled pedestrians/persons and cyclists/bikers as the other two dynamic categories. Apart from the moving objects, the remaining pixels can be regarded as the static background. 

\begin{table*}[t!]
  \centering
\caption{\textbf{Depth Evaluation of Background and Dynamic Objects.} }
\label{tab:back_dyn_evalu}
\resizebox{0.92\textwidth}{!}
{
\begin{threeparttable}
\footnotesize
\begin{tabular}{c||l||c|c|c|c|c|c|c}
  \toprule
    
 \multirow{2}{*}{\begin{tabular}{@{}c@{}} Region \\ Amount, Percent \end{tabular}} & \multirow{2}{*}{Method}
 & \multicolumn{4}{c|}{Error metric $\downarrow	$} & \multicolumn{3}{c}{Accuracy metric $\uparrow  $}\\
 \cline{3-6}
  \cline{7-9}
   &   & Abs Rel & Sq Rel & RMSE  & RMSE log & $\delta < 1.25 $ & $\delta < 1.25^{2}$ & $\delta < 1.25^{3}$   \\
  \hline
  \hline
 \multirow{5}{*}{\begin{tabular}{@{}c@{}}Background \\ 11379k, 96.29\% \end{tabular}} &
DORN~\cite{fu2018deep} 
&   0.071  &   0.299  &   2.741  &   0.117  &   0.933  &   0.985  &   0.995  \\
\clinegraytwo
& Monodepth2-S~\cite{godard2019digging} 
&   0.107  &   0.823  &   4.844  &   0.200  &   0.869  &   0.951  &   0.977  \\
\clinegraytwo
& Monodepth2-M~\cite{godard2019digging}
&   0.110  &   0.815  &   4.697  &   0.183  &   0.883  &   0.963  &   0.983  \\
\clinegraytwo
& w/o outlier masking 
&   0.110  &   0.809  &   4.668  &   0.183  &   0.883  &   0.962  &   0.983  \\
& DiPE 
&   0.107  &   0.784  &   4.614  &   0.180  &   0.886  &   0.964  &   0.984  \\
\hline
\hline
 
\multirow{5}{*}{\begin{tabular}{@{}c@{}c@{}}Dynamic \\ Objects \\ 438k, 3.71\% \end{tabular}} &
DORN~\cite{fu2018deep} 
&   0.099  &   0.444  &   2.605  &   0.151  &   0.884  &   0.960  &   0.986  \\
\clinegraytwo
& Monodepth2-S~\cite{godard2019digging} 
&   0.167  &   2.064  &   6.487  &   0.320  &   0.780  &   0.888  &   0.930  \\
\clinegraytwo
& Monodepth2-M~\cite{godard2019digging}
&   0.219  &   3.052  &   7.065  &   0.320  &   0.731  &   0.886  &   0.933  \\
\clinegraytwo
& w/o outlier masking 
&   0.220  &   3.173  &   7.338  &   0.330  &   0.719  &   0.872  &   0.928  \\
& DiPE 
&   0.215  &   3.083  &   7.172  &   0.319  &   0.737  &   0.883  &   0.931  \\

\hline
\hline

\multirow{5}{*}{\begin{tabular}{@{}c@{}c@{}c@{}} Similarly \\ Moving \\ Vehicles \\ 64k, 0.54\% \end{tabular}} &
DORN~\cite{fu2018deep} 
&   0.098  &   0.477  &   3.121  &   0.162  &   0.877  &   0.954  &   0.983  \\
\clinegraytwo
& Monodepth2-S~\cite{godard2019digging} 
&   0.147  &   2.633  &   7.997  &   0.300  &   0.851  &   0.892  &   0.914  \\
\clinegraytwo
& Monodepth2-M~\cite{godard2019digging}
&   0.264  &   4.966  &  10.173  &   0.361  &   0.672  &   0.850  &   0.903  \\
\clinegraytwo
& w/o outlier masking 
&   0.311  &   6.902  &  11.648  &   0.403  &   0.623  &   0.801  &   0.872  \\
& DiPE 
&   0.278  &   6.076  &  10.956  &   0.376  &   0.677  &   0.828  &   0.888  \\

\hline
\hline

\multirow{5}{*}{\begin{tabular}{@{}c@{}c@{}c@{}} Dissimilarly \\ Moving \\ Vehicles \\ 50k, 0.42\% \end{tabular}} &
DORN~\cite{fu2018deep} 
&   0.118  &   0.774  &   4.024  &   0.192  &   0.834  &   0.937  &   0.972  \\
\clinegraytwo
& Monodepth2-S~\cite{godard2019digging} 
&   0.158  &   1.931  &   7.339  &   0.326  &   0.816  &   0.899  &   0.923  \\
\clinegraytwo
& Monodepth2-M~\cite{godard2019digging}
&   0.186  &   2.455  &   7.652  &   0.330  &   0.728  &   0.877  &   0.926  \\
\clinegraytwo
& w/o outlier masking 
&   0.190  &   2.554  &   7.922  &   0.343  &   0.720  &   0.844  &   0.919  \\
& DiPE 
&   0.174  &   2.228  &   7.483  &   0.325  &   0.749  &   0.886  &   0.925  \\

\hline
\hline
\multirow{5}{*}{\begin{tabular}{@{}c@{}c@{}c@{}} Slowly \\ Moving \\ Vehicles \\ 216k, 1.82\% \end{tabular}} &
DORN~\cite{fu2018deep} 
&   0.086  &   0.308  &   1.898  &   0.131  &   0.909  &   0.965  &   0.988  \\
\clinegraytwo
& Monodepth2-S~\cite{godard2019digging} 
&   0.144  &   1.117  &   4.779  &   0.305  &   0.789  &   0.901  &   0.941  \\
\clinegraytwo
& Monodepth2-M~\cite{godard2019digging}
&   0.172  &   1.619  &   5.125  &   0.285  &   0.795  &   0.920  &   0.950  \\
\clinegraytwo
& w/o outlier masking 
&   0.167  &   1.483  &   5.038  &   0.289  &   0.784  &   0.912  &   0.950  \\
& DiPE 
&   0.167  &   1.536  &   4.960  &   0.278  &   0.792  &   0.919  &   0.950  \\

\hline
\hline

\multirow{5}{*}{\begin{tabular}{@{}c@{}c@{}} Pedestrians \\ or Persons \\ 67k, 0.57\% \end{tabular}} &
DORN~\cite{fu2018deep} 
&   0.136  &   0.775  &   3.289  &   0.187  &   0.828  &   0.953  &   0.985  \\
\clinegraytwo
& Monodepth2-S~\cite{godard2019digging} 
&   0.260  &   4.718  &   8.661  &   0.384  &   0.683  &   0.827  &   0.900  \\
\clinegraytwo
& Monodepth2-M~\cite{godard2019digging}
&   0.347  &   6.635  &   9.070  &   0.393  &   0.608  &   0.816  &   0.897  \\
\clinegraytwo
& w/o outlier masking 
&   0.322  &   5.824  &   9.052  &   0.386  &   0.619  &   0.818  &   0.902  \\
& DiPE 
&   0.329  &   6.165  &   8.785  &   0.384  &   0.631  &   0.817  &   0.902  \\

\hline
\hline

\multirow{5}{*}{\begin{tabular}{@{}c@{}c@{}} Cyclists \\ or Bikers \\ 42k, 0.35\%  \end{tabular}} &
DORN~\cite{fu2018deep} 
&   0.118  &   0.409  &   2.198  &   0.146  &   0.877  &   0.983  &   0.996  \\
\clinegraytwo
& Monodepth2-S~\cite{godard2019digging} 
&   0.183  &   2.022  &   5.413  &   0.275  &   0.733  &   0.894  &   0.951  \\
\clinegraytwo
& Monodepth2-M~\cite{godard2019digging}
&   0.225  &   2.518  &   5.472  &   0.281  &   0.697  &   0.888  &   0.959  \\
\clinegraytwo
& w/o outlier masking
&   0.226  &   2.683  &   5.678  &   0.280  &   0.687  &   0.896  &   0.953  \\
 & DiPE 
&   0.232  &   2.581  &   5.561  &   0.281  &   0.700  &   0.877  &   0.948  \\
\bottomrule
\end{tabular}

\begin{tablenotes}[flushleft]
\item Note: Monodepth2-S: the stereo-image-training; Monodepth2-M: the monocular-video-training.
\end{tablenotes}
\end{threeparttable}

}
\vspace{-9pt}
\end{table*}

In evaluating the depth prediction results for the region belonging to a particular category, we cannot simply average the results of all test images as in typical depth evaluation. Because the numbers of pixels in different test images are different, and the test images may contain no pixel for any moving pattern. Therefore, we average the results from different test samples by weighting their number of pixels belonging to a specific moving pattern. We evaluate the five different categories, their union as dynamic objects and the static background for the supervised DORN~\cite{fu2018deep}, the stereo image based unsupervised Monodepth2-S \cite{godard2019digging} and three monocular video based unsupervised variants Monodepth2-M \cite{godard2019digging}, DiPE without outlier masking and DiPE. In calculating the aligning scale factor for the monocular video based unsupervised methods, we use the median value of pixels only from the background instead of the whole image. The value from dynamic pixels is much less reliable. Tab.~\ref{tab:back_dyn_evalu} shows the evaluation results, where we can find many interesting observations.

The first column of Tab.~\ref{tab:back_dyn_evalu} shows the amount and percent of pixels with ground-truth depth for different categories. The static background dominates the scenes by accounting for over 96\%, while the dynamic objects only take up 3.71\%. Therefore, the overall performance of conventional depth evaluation largely depends on the static background. However, the depth estimation accuracy for dynamic objects is much more important for applications, such as autonomous driving or robot navigation. That is the reason why we propose to do depth evaluation for dynamic objects separately.

Comparing the metrics between the dynamic objects with the background shows that all methods' performance on dynamic objects exhibits a larger degradation. The extent of degradation of different kinds of methods is different. The depth-supervised DORN~\cite{fu2018deep} has the smallest degradation, with the Abs Rel error increasing by 40\% and the $\delta < 1.25 $ accuracy having a decrease of 0.05. The stereo image based unsupervised method Monodepth2-S \cite{godard2019digging} shows a bigger degeneration, with the Abs Rel error increasing by 56\% times and the $\delta < 1.25 $ accuracy having a decrease of 0.09. In contrast, the monocular video based unsupervised variants have a much more significant degradation. To be specific, the Abs Rel error doubles, and the $\delta < 1.25 $ accuracy has a drop of about 0.15. The degradation of DORN and Monodepth2-S indicates that the depth of dynamics objects is more challenging to learn than that of the static background, even though this method is not sensitive to dynamics in theory.

On the other hand, the vulnerability to dynamic objects of the monocular video based unsupervised training methods can further {aggravate} the degradation at a similar level, even when Monodepth2 and this paper have proposed some masking techniques to alleviate the influence of dynamic objects. We are the first to quantitatively evaluate the performance degradation on dynamic objects for monocular video based unsupervised methods. The performance gap between the monocular video based unsupervised methods and the stereo image based unsupervised or even the directly supervised methods is still large, which means enough room for improvement in the near future. On the other hand, we believe this gap may also be the limit of improvement on this front. We might need to look at other sources for further improvement.

When comparing the performance of different kinds of moving vehicles, we can easily find that the performance on slowly moving vehicles is the best one for all methods, which indicates the depth of shower moving vehicles is easier to learn. The supervised DORN \cite{fu2018deep} and stereo image based Monodepth2-S \cite{godard2019digging} perform worst in dissimilarly moving vehicles among all the dynamic patterns, but the gap to similarly moving vehicles is small. However, the monocular video based methods perform the worst on the similarly moving vehicles, and the gap with dissimilarly moving vehicles is considerable. This can be explained by Fig.~\ref{fig:om}, which indicates that in the same movement, the depth error caused by similarly moving objects is much bigger than that of dissimilarly moving objects. That is the reason why the problem of such a moving pattern was easily identified and tackled \cite{casser2019depth, luo2019every, godard2019digging}.

Interestingly, the slowest moving pedestrians/persons show the highest inaccuracy. Their Abs Rel errors are approximately 1.5 times that of overall dynamic objects for all methods. On the other hand, the decrease in the $\delta < 1.25 $ accuracy is different for different methods. While the directly supervised DORN \cite{fu2018deep} has a decrease of about 0.05, the unsupervised ones have a decrease of around 0.1. We conjecture that it is the much higher appearance  diversity of pedestrians/persons than the vehicles that make them very difficult to learn. {For a similar reason, the cyclists/bikers, though slower but diverse, show worse performance than the vehicles for all methods.}

To find the effectiveness of our approach to different categories of regions, we can compare the performance of three monocular video 
based unsupervised variants. When looking at their performance on the static background, we can find that DiPE improves the performance by a decrease of 0.03 in the Abs Rel error and an increase of 0.3\% in the $\delta < 1.25 $ accuracy comparing with Monodepth2-M \cite{godard2019digging} and when turning off the outlier masking technique. Obviously, the improvement on static background is almost the same as that for all regions from Table.~\ref{tab:kitti_eigen} and Table.~\ref{tab:kitti_eigen_ablation}. Considering that the static background accounts for over 96\% of the area in the test set, the overall improvement is mainly driven by the improvement in the static background. Although the outlier masking technique is designed to avoid the erroneous signal from moving objects, especially the dissimilarly moving objects, it turns out to also help to improve accuracy in static background regions too. 
We conjecture that ignoring the higher photometric errors mainly from moving objects is helpful to the optimization of the smaller errors from the static background, thus making the learning of background better.

{When it comes to the general dynamic objects, DiPE has larger improvement upon the baseline without the outlier masking technique}, i.e., a decrease of 0.05 in the Abs Rel error and an increase of 1.8\% in the $\delta < 1.25 $ accuracy increases. This verifies the effectiveness of the outlier masking in improving the depth estimation of dynamic objects. Although DiPE can still have an improvement of 0.6\% in the $\delta < 1.25 $ accuracy upon Monodepth2-M \cite{godard2019digging}, it outperforms Monodepth2-M only in some metrics. This also happens for similarly moving vehicles. Because there are different choices on multi-scale methods between DiPE variants and Monodepth2-M, we speculate that multi-scale methods also impact the depth learning of the dynamic objects.

For the dissimilarly moving vehicles, DiPE has much more significant improvement upon the other two variants than than it does on the other categories of regions. To be specific, there is a reduction of 0.1 in the Abs Rel error and an enhancement of 2\% in the $\delta < 1.25 $ accuracy. This demonstrates that the proposed outlier masking has a more significant impact on this particular kind of moving pattern. In the last three categories of moving objects, DiPE still has different extents of improvements upon the other two variants.

\begin{table}[ht]
\centering
\caption{
\textbf{Visual odometry results on KITTI odometry split~\cite{geiger2013vision}.} }
\resizebox{0.95\columnwidth}{!}{
\begin{threeparttable}

\begin{tabular}{l|c|c|c}
 \toprule
Method & \textbf{Sequence 09}     & \textbf{Sequence 10} & \textbf{\# frames} \\ \hline
\multicolumn{1}{l|}{ORB-SLAM \cite{mur2015orb}} & 0.014$\pm$0.008 & 0.012$\pm$0.011 & -  \\ \hline \hline
\multicolumn{1}{l|}{SfMLearner \cite{zhou2017unsupervised}} & 0.021$\pm$0.017 & 0.020$\pm$0.015 & 5\\ 
\multicolumn{1}{l|}{DF-Net \cite{zou2018df}} & 0.017$\pm$0.007 & 0.015$\pm$0.009 & 5 \\
\multicolumn{1}{l|}{SC-SfMLearner~\cite{bian2019depth}} & 0.016$\pm$0.007 & 0.015$\pm$0.015 & 5 \\
\multicolumn{1}{l|}{GeoNet \cite{yin2018geonet}} & {0.012}$\pm$0.007 & {0.012$\pm$0.009} & 5 \\  
\multicolumn{1}{l|}{\textbf{DiPE} (Ours)} &\textbf{0.012$\pm$0.006} &\textbf{0.012$\pm$0.008}& 5 \\ 
\hline
\hline
\multicolumn{1}{l|}{DDVO \cite{wang2018learning}} & 0.045$\pm$0.108 & 0.033$\pm$0.074 & 3 \\ 
\multicolumn{1}{l|}{Vid2Depth \cite{mahjourian2018unsupervised}} & 0.013$\pm$0.010 & {0.012}$\pm$0.011 & 3 \\ 
\multicolumn{1}{l|}{EPC++ \cite{luo2019every}} & 0.013$\pm$0.007 & \textbf{0.012$\pm${0.008}} & 3\\ 
%\multicolumn{1}{l|}{Struct2Depth \cite{casser2019depth}} & 0.011$\pm$0.006 & 0.011$\pm$0.010 & 3 \\ 
\multicolumn{1}{l|}{\textbf{DiPE} (Ours)} & \textbf{0.012$\pm$0.006} & \textbf{0.012$\pm$0.008} & 3 \\ 
\hline 
\hline
\multicolumn{1}{l|}{{Monodepth2 \cite{godard2019digging}}} & 0.017$\pm$0.008 & 0.015$\pm$0.010 & 2 \\ 
%\multicolumn{1}{l|}{{Gordon \etal \cite{Gordon_2019_ICCV}}} & 0.012$\pm$0.016 & 0.010$\pm$0.010 & 2 \\ 
\multicolumn{1}{l|}{\textbf{DiPE} (Ours)} & \textbf{0.013$\pm$0.006} & \textbf{0.012$\pm$0.008}  & 2 \\ 
\bottomrule
\end{tabular}

\begin{tablenotes}[flushleft]
\footnotesize   
\item Note: Results show the average absolute trajectory error and standard deviation, in meters.
\end{tablenotes}
\end{threeparttable}
}
\label{tab:odom}
\vspace{-10pt}
\end{table}

\subsection{KITTI Odometry Benchmark}
To show DiPE's effectiveness on ego-motion estimation, we also experiment on the official odometry split of the KITTI dataset \cite{geiger2013vision}. We adopt three different input settings for the ego-motion network. The number of frames is set at $2$, $3$, and $5$, respectively. When training the Pose CNN with input as $2$ or $3$ frames, we build DiPE on the second baseline model. However, when experimenting with the 5-frame-input Pose CNN, we do not adopt the minimum reprojection technique because it almost masks out all the pixels from the source views with indexes of $t-2$ and $t+2$, and the motion estimation for these two views is inferior.

\begin{table*}[t!]
\centering
\caption{\textbf{Quantitative Results of experiments on Cityscapes.}}  
\label{tab:cs_eval} 
\resizebox{0.98\textwidth}{!}
{
\begin{threeparttable}
\footnotesize
\begin{tabular}{c|l|c|c|c|c|c|c|c}
  \toprule
    
 \multirow{2}{*}{\begin{tabular}{@{}c@{}} Region \\ Amount, Percent \end{tabular}} & \multirow{2}{*}{Methods}
 & \multicolumn{4}{c|}{Error Metrics $\downarrow	$} & \multicolumn{3}{c}{Accuracy Metrics $\uparrow  $}\\
 \cline{3-6}
  \cline{7-9}
   &   & Rel & SqRel & RMS  & RMS$_{log}$ & $\delta < 1.25 $ & $\delta < 1.25^{2}$ & $\delta < 1.25^{3}$   \\

   \hline
 
\multirow{5}{*}{\begin{tabular}{@{}c@{}c@{}} Background \\ 531093k, 88.57\% \end{tabular}} 
& Monodepth2~\cite{godard2019digging}
&   0.156  &   2.390  &   8.149  &   0.223  &   0.804  &   0.947  &   0.980  \\
\clinegraytwo
& Baseline II  
&   0.167  &   2.668  &   8.468  &   0.235  &   0.786  &   0.938  &   0.976  \\
& DiPE w/o weighted multi-scale 
&   0.162  &   2.598  &   8.381  &   0.228  &   0.797  &   0.941  &   0.978  \\
& DiPE w/o outlier masking 
&   0.157  &   2.384  &   8.193  &   0.225  &   0.800  &   0.945  &   0.979  \\
& DiPE 
&   0.155  &   2.381  &   8.127  &   0.220  &   0.808  &   0.947  &   0.980  \\
\hline

\multirow{5}{*}{\begin{tabular}{@{}c@{}c@{}c@{}} Objects \\ 68510k, 11.43\% \end{tabular}} 
& Monodepth2~\cite{godard2019digging}
&   0.383  &  13.864  &  10.406  &   0.373  &   0.676  &   0.825  &   0.894  \\
\clinegraytwo
& Baseline II  
&   0.398  &  13.801  &  10.944  &   0.407  &   0.637  &   0.802  &   0.875  \\
& DiPE w/o weighted multi-scale 
&   0.409  &  15.187  &  10.717  &   0.382  &   0.669  &   0.825  &   0.890  \\
& DiPE w/o outlier masking 
&   0.394  &  14.473  &  10.607  &   0.384  &   0.664  &   0.817  &   0.883  \\
& DiPE 
&   0.365  &  13.401  &   9.742  &   0.336  &   0.697  &   0.861  &   0.924  \\
\bottomrule
\end{tabular}
\begin{tablenotes}[flushleft]
\footnotesize              
\item Note: All methods are trained with monocular videos. Baseline II is the same as the one of Tab.~\ref{tab:kitti_eigen_ablation}, \ie, DiPE without our two contributions.
\end{tablenotes}
\end{threeparttable}
}
\end{table*}

In evaluation, we adopt the conventional metric proposed by Zhou \etal~\cite{zhou2017unsupervised}, i.e., the Absolute Trajectory Error (ATE)~\cite{mur2015orb} in $5$-frame snippets. Tab.~\ref{tab:odom} presents the results. The results of other models are taken from their corresponding papers. Among models with the three different input settings, DiPE achieves the best performance. Notably, in the pair-input ego-motion network setting, DiPE significantly outperforms Monodepth2 \cite{godard2019digging}. There is no significant performance difference among different motion network settings for DiPE; thus, DiPE is robust to different motion network input settings.

\subsection{Experiments on Cityscapes}

{
To examine the generalization ability of our proposed method, we further perform experiments on another popular driving dataset,  CityScapes~\cite{Cordts2016Cityscapes}. The dataset contains onboard videos, thus we can use it on unsupervised depth training. The dataset also provides the disparity maps computed by Semi-Global Matching (SGM)~\cite{hirschmuller2007stereo} algorithm on the stereo images. As the baseline of stereo images is also given, we can convert the disparity into depth for evaluation. 
}

{
Cityscapes is originally used in semantic segmentation, which provides 
2975 samples from 18 cities for training, 500 samples from 3 cities for validation, and 1525 images from 6 cities for testing.  
We use videos from the 18 cities of the training set to construct the training set, resulting in 71400 samples. 
Considering that the fine annotations for the validation set are publicly available but those of the test set are not, we use the original test set as the validation set, and the original validation set as the test set in our experiments. By doing that, we can fully utilize the semantic annotations for performing depth evaluation of probably moving objects like~\ref{sec:dynamic}. 
To be specific, we use the regions from the segmentations of ``person", ``rider", ``car", ``truck", ``bus", ``caravan", ``trailer", ``train", ``motorcycle" and ``bicycle" as the masks of foreground object to perform such evaluation.  
}

\begin{figure}[t]
\centering
\resizebox{1.01\linewidth}{!}{
\newcommand{\shiftleft}[2]{\makebox[-6pt][r]{\makebox[#1][l]{#2}}}
\newcommand{\imlabel}[2]{\includegraphics[width=0.49\columnwidth]{#1}%
\raisebox{36pt}{\shiftleft{56pt}{\makebox[-2pt][r]{\footnotesize #2}} }}

\centering
\renewcommand{\arraystretch}{0.5}
\begin{tabular}{@{\hskip -4.5mm}c@{\hskip 1.5mm}c@{\hskip 1.5mm}c}
\imlabel{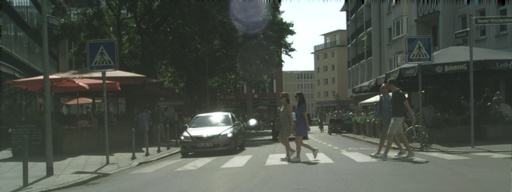}{} &
\imlabel{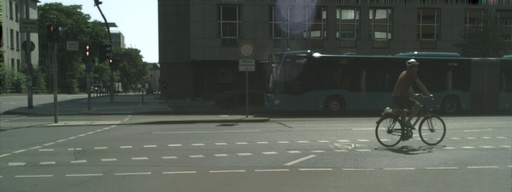}{} &
\imlabel{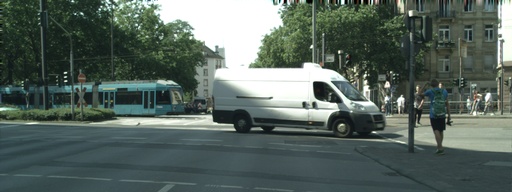}{} \\
\imlabel{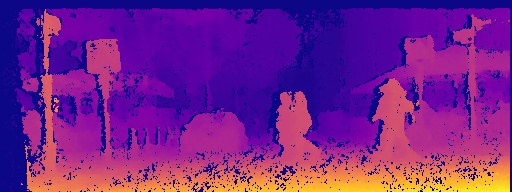}{\textcolor{white}{Ground Truth \quad}} &
\imlabel{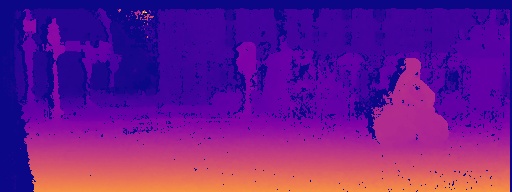}{\textcolor{white}{Ground Truth \quad}} &
\imlabel{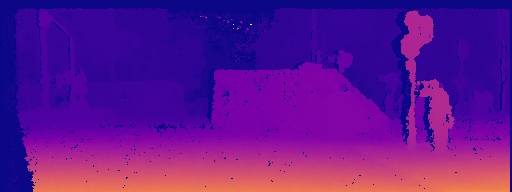}{\textcolor{white}{Ground Truth \quad}} \\
\imlabel{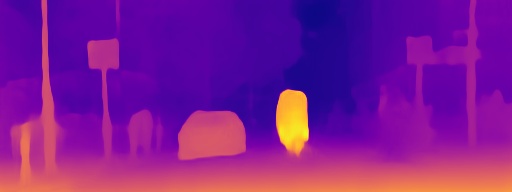}{\textcolor{white}{Monodepth2~\cite{godard2019digging}}} &
\imlabel{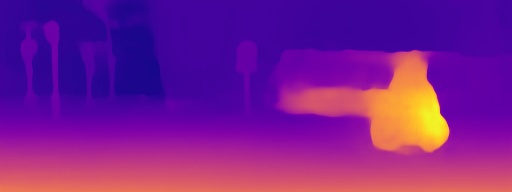}{\textcolor{white}{Monodepth2~\cite{godard2019digging}}} &
\imlabel{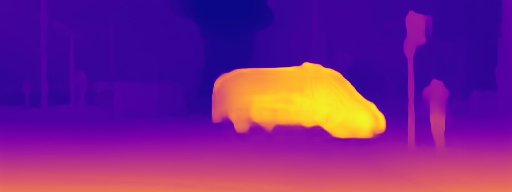}{\textcolor{white}{Monodepth2~\cite{godard2019digging}}} \\
\imlabel{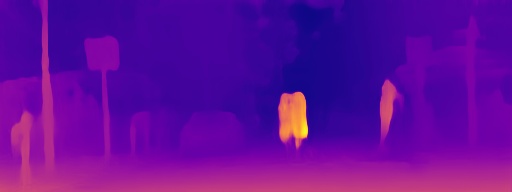}{\textcolor{white}{Baseline II \ \qquad}} &
\imlabel{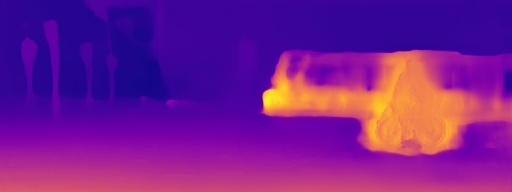}{\textcolor{white}{Baseline II \ \qquad}} &
\imlabel{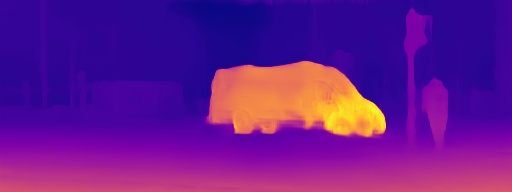}{\textcolor{white}{Baseline II \ \qquad}} \\
\imlabel{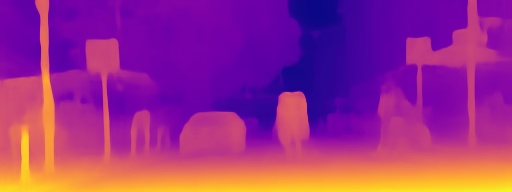}{\textcolor{white}{DiPE  \ \ \qquad \qquad }} &  
\imlabel{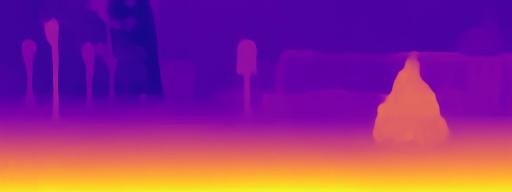}{\textcolor{white}{DiPE  \ \ \qquad \qquad }} &
\imlabel{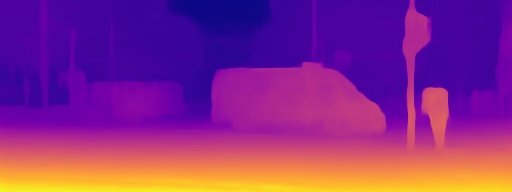}{\textcolor{white}{DiPE  \ \ \qquad \qquad }}\\

\end{tabular}

}
\caption{{\bf Qualitative comparison on Cityscapes.} DiPE handles the moving objects better than Monodepth2~\cite{godard2019digging} and the baseline.}
\label{fig:cityscapes}
\vspace{-8pt}
\end{figure}

{
The training settings are almost the same as KITTI in Sec.~\ref{sc:impl} except the input size. As the images in Cityscapes cover the head of the car, we first remove the lower 256 lines of pixels on the bottom following SfMlearner~\cite{zhou2017unsupervised}. The cropped image size is  $768\times2048$. To obtain a close size with KITTI ($192\times640$), we further resize the cropped images into} {($192\times512$) for the networks. In testing, we also up-sample the predicted depth maps to $768\times2048$ before evaluation. 
}

{
The quantitative results are shown on Tab.~\ref{tab:cs_eval}. The organization Tab.~\ref{tab:cs_eval} is like Tab.~\ref{tab:back_dyn_evalu}. 
The object pixels here take a larger proportion than those of the moving objects in KITTI. This is due to that the data of Cityscapes is captured in urban street scenes where there are many objects and the objects may contain some static ones. Compared with Monodepth2~\cite{godard2019digging}, DiPE is only slightly better on the background but remarkably outperforms it on the objects. Similar phenomena can be observed when our DiPE and the variants without our technical contributions. The results further verify our proposed methods are much more useful for objects. The qualitative results in Fig.~\ref{fig:cityscapes} demonstrate the effectiveness of our method.}

\section{CONCLUSION}
In this paper, we have demonstrated that carefully processing the photometric errors for unsupervised learning of depth and ego-motion from monocular videos can significantly solve the intrinsic difficulties, \ie, occlusion, and scene dynamics. 
We have introduced the outlier masking technique to exclude the irregular photometric errors that may mislead the network learning. This technique is useful to deal with occlusion and scene dynamics, especially for contra-directionally moving objects. Moreover, we have proposed an efficient and effective weighted multi-scale scheme to avoid the artifacts brought by multi-scale training. 
Unlike other methods that introduce extra modules, our approach is simple, as they can be conveniently incorporated into the unsupervised geometry learning framework. 
We have experimentally proven the effectiveness of our two contributions and built the state-of-the-art model, DiPE, on both monocular depth and ego-motion estimation. Furthermore, we are the first to evaluate the depth estimation results with different motion patterns. The evaluation further verifies
the effectiveness of our approach. Comparing the performance of different kinds of methods on different motion properties, we point out the possible room to improve the moving objects for the unsupervised monocular training methods.

% or


\ifCLASSOPTIONcaptionsoff
  \newpage
\fi



%\bibliographystyle{IEEEtran}

{\small
  \bibliographystyle{IEEEtran}
  \bibliography{ref}

\begin{thebibliography}{10}
\providecommand{\url}[1]{#1}
\csname url@samestyle\endcsname
\providecommand{\newblock}{\relax}
\providecommand{\bibinfo}[2]{#2}
\providecommand{\BIBentrySTDinterwordspacing}{\spaceskip=0pt\relax}
\providecommand{\BIBentryALTinterwordstretchfactor}{4}
\providecommand{\BIBentryALTinterwordspacing}{\spaceskip=\fontdimen2\font plus
\BIBentryALTinterwordstretchfactor\fontdimen3\font minus
  \fontdimen4\font\relax}
\providecommand{\BIBforeignlanguage}[2]{{%
\expandafter\ifx\csname l@#1\endcsname\relax
\typeout{** WARNING: IEEEtran.bst: No hyphenation pattern has been}%
\typeout{** loaded for the language `#1'. Using the pattern for}%
\typeout{** the default language instead.}%
\else
\language=\csname l@#1\endcsname
\fi
#2}}
\providecommand{\BIBdecl}{\relax}
\BIBdecl

\bibitem{poggi2019confidence}
M.~Poggi, G.~Agresti, F.~Tosi, P.~Zanuttigh, and S.~Mattoccia, ``Confidence
  estimation for tof and stereo sensors and its application to depth data
  fusion,'' \emph{IEEE Sensors Journal}, vol.~20, no.~3, pp. 1411--1421, 2019.

\bibitem{zhao2020fusion}
X.~Zhao, P.~Sun, Z.~Xu, H.~Min, and H.~Yu, ``Fusion of 3d lidar and camera data
  for object detection in autonomous vehicle applications,'' \emph{IEEE Sensors
  Journal}, vol.~20, no.~9, pp. 4901--4913, 2020.

\bibitem{mur2015orb}
R.~Mur-Artal, J.~M.~M. Montiel, and J.~D. Tardos, ``Orb-slam: a versatile and
  accurate monocular slam system,'' \emph{IEEE transactions on robotics},
  vol.~31, no.~5, pp. 1147--1163, 2015.

\bibitem{engel2014lsd}
J.~Engel, T.~Sch{\"o}ps, and D.~Cremers, ``Lsd-slam: Large-scale direct
  monocular slam,'' in \emph{European conference on computer vision}.\hskip 1em
  plus 0.5em minus 0.4em\relax Springer, 2014, pp. 834--849.

\bibitem{schoenberger2016sfm}
J.~L. Sch\"{o}nberger and J.-M. Frahm, ``Structure-from-motion revisited,'' in
  \emph{Conference on Computer Vision and Pattern Recognition}, 2016.

\bibitem{eigen2014depth}
D.~Eigen, C.~Puhrsch, and R.~Fergus, ``Depth map prediction from a single image
  using a multi-scale deep network,'' in \emph{Advances in neural information
  processing systems}, 2014, pp. 2366--2374.

\bibitem{wang2015towards}
P.~Wang, X.~Shen, Z.~Lin, S.~Cohen, B.~Price, and A.~L. Yuille, ``Towards
  unified depth and semantic prediction from a single image,'' in
  \emph{Proceedings of the IEEE Conference on Computer Vision and Pattern
  Recognition}, 2015, pp. 2800--2809.

\bibitem{liu2015learning}
F.~Liu, C.~Shen, G.~Lin, and I.~Reid, ``Learning depth from single monocular
  images using deep convolutional neural fields,'' \emph{IEEE transactions on
  pattern analysis and machine intelligence}, vol.~38, no.~10, pp. 2024--2039,
  2015.

\bibitem{laina2016deeper}
I.~Laina, C.~Rupprecht, V.~Belagiannis, F.~Tombari, and N.~Navab, ``Deeper
  depth prediction with fully convolutional residual networks,'' in \emph{3D
  Vision (3DV), 2016 Fourth International Conference on}.\hskip 1em plus 0.5em
  minus 0.4em\relax IEEE, 2016, pp. 239--248.

\bibitem{xu2017multi}
D.~Xu, E.~Ricci, W.~Ouyang, X.~Wang, and N.~Sebe, ``Multi-scale continuous crfs
  as sequential deep networks for monocular depth estimation,'' in
  \emph{Proceedings of the IEEE Conference on Computer Vision and Pattern
  Recognition}, 2017.

\bibitem{zeng2017geocuedepth}
Y.~Zeng, Y.~Hu, S.~Liu, Q.~Tang, J.~Ye, and X.~Li, ``Geocuedepth: exploiting
  geometric structure cues to estimate depth from a single image,'' in
  \emph{2017 IEEE/RSJ International Conference on Intelligent Robots and
  Systems}, 2017.

\bibitem{cao2017estimating}
Y.~Cao, Z.~Wu, and C.~Shen, ``Estimating depth from monocular images as
  classification using deep fully convolutional residual networks,'' \emph{IEEE
  Transactions on Circuits and Systems for Video Technology}, vol.~28, no.~11,
  pp. 3174--3182, 2017.

\bibitem{li2018monocular}
B.~Li, Y.~Dai, and M.~He, ``Monocular depth estimation with hierarchical fusion
  of dilated cnns and soft-weighted-sum inference,'' \emph{Pattern
  Recognition}, vol.~83, pp. 328--339, 2018.

\bibitem{fu2018deep}
H.~Fu, M.~Gong, C.~Wang, K.~Batmanghelich, and D.~Tao, ``Deep ordinal
  regression network for monocular depth estimation,'' in \emph{IEEE Conference
  on Computer Vision and Pattern Recognition}, 2018.

\bibitem{liu2016vehicle}
C.~Liu, R.~Fujishiro, L.~Christopher, and J.~Zheng, ``Vehicle--bicyclist
  dynamic position extracted from naturalistic driving videos,'' \emph{IEEE
  transactions on intelligent transportation systems}, vol.~18, no.~4, pp.
  734--742, 2016.

\bibitem{zhou2017unsupervised}
T.~Zhou, M.~Brown, N.~Snavely, and D.~G. Lowe, ``Unsupervised learning of depth
  and ego-motion from video,'' in \emph{Proceedings of the IEEE Conference on
  Computer Vision and Pattern Recognition}, 2017, pp. 1851--1858.

\bibitem{klodt2018supervising}
M.~Klodt and A.~Vedaldi, ``Supervising the new with the old: learning sfm from
  sfm,'' in \emph{Proceedings of the European Conference on Computer Vision},
  2018, pp. 698--713.

\bibitem{bian2019depth}
J.-W. Bian, Z.~Li, N.~Wang, H.~Zhan, C.~Shen, M.-M. Cheng, and I.~Reid,
  ``Unsupervised scale-consistent depth and ego-motion learning from monocular
  video,'' in \emph{Thirty-third Conference on Neural Information Processing
  Systems}, 2019.

\bibitem{godard2019digging}
C.~Godard, O.~Mac~Aodha, M.~Firman, and G.~J. Brostow, ``Digging into
  self-supervised monocular depth estimation,'' in \emph{Proceedings of the
  IEEE International Conference on Computer Vision}, 2019, pp. 3828--3838.

\bibitem{wang2020unsupervised}
G.~Wang, C.~Zhang, H.~Wang, J.~Wang, Y.~Wang, and X.~Wang, ``Unsupervised
  learning of depth, optical flow and pose with occlusion from 3d geometry,''
  \emph{IEEE Transactions on Intelligent Transportation Systems}, pp. 1--13,
  2020.

\bibitem{luo2019every}
C.~Luo, Z.~Yang, P.~Wang, Y.~Wang, W.~Xu, R.~Nevatia, and A.~Yuille, ``Every
  pixel counts++: Joint learning of geometry and motion with 3d holistic
  understanding,'' \emph{IEEE transactions on pattern analysis and machine
  intelligence}, vol.~42, no.~10, pp. 2624--2641, 2019.

\bibitem{yin2018geonet}
Z.~Yin and J.~Shi, ``Geonet: Unsupervised learning of dense depth, optical flow
  and camera pose,'' in \emph{The IEEE Conference on Computer Vision and
  Pattern Recognition}, 2018.

\bibitem{zou2018df}
Y.~Zou, Z.~Luo, and J.-B. Huang, ``Df-net: Unsupervised joint learning of depth
  and flow using cross-task consistency,'' in \emph{Proceedings of the European
  Conference on Computer Vision}, 2018.

\bibitem{vijayanarasimhan2017sfm}
S.~Vijayanarasimhan, S.~Ricco, C.~Schmid, R.~Sukthankar, and K.~Fragkiadaki,
  ``Sfm-net: Learning of structure and motion from video,'' \emph{arXiv
  preprint arXiv:1704.07804}, 2017.

\bibitem{casser2019depth}
V.~Casser, S.~Pirk, R.~Mahjourian, and A.~Angelova, ``Depth prediction without
  the sensors: Leveraging structure for unsupervised learning from monocular
  videos,'' in \emph{Proceedings of the AAAI Conference on Artificial
  Intelligence}, 2019.

\bibitem{Gordon_2019_ICCV}
A.~Gordon, H.~Li, R.~Jonschkowski, and A.~Angelova, ``Depth from videos in the
  wild: Unsupervised monocular depth learning from unknown cameras,'' in
  \emph{The IEEE International Conference on Computer Vision}, October 2019.

\bibitem{yuan2020independent}
J.~Yuan, G.~Zhang, F.~Li, J.~Liu, L.~Xu, S.~Wu, T.~Jiang, D.~Guo, and Y.~Xie,
  ``Independent moving object detection based on a vehicle mounted binocular
  camera,'' \emph{IEEE Sensors Journal}, 2020.

\bibitem{jiang2020dipe}
H.~Jiang, L.~Ding, Z.~Sun, and R.~Huang, ``Dipe: Deeper into photometric errors
  for unsupervised learning of depth and ego-motion from monocular videos,'' in
  \emph{In IEEE/RSJ International Conference on Intelligent Robots and
  Systems}, 2020.

\bibitem{kittidepthserver}
``Kitti single depth evaluation server.''
  \url{http://www.cvlibs.net/datasets/kitti/eval_depth.php?benchmark=depth_prediction},
  2017.

\bibitem{Cordts2016Cityscapes}
M.~Cordts, M.~Omran, S.~Ramos, T.~Rehfeld, M.~Enzweiler, R.~Benenson,
  U.~Franke, S.~Roth, and B.~Schiele, ``The cityscapes dataset for semantic
  urban scene understanding,'' in \emph{Proc. of the IEEE Conference on
  Computer Vision and Pattern Recognition (CVPR)}, 2016.

\bibitem{godard2017unsupervised}
C.~Godard, O.~Mac~Aodha, and G.~J. Brostow, ``Unsupervised monocular depth
  estimation with left-right consistency,'' in \emph{Proceedings of the IEEE
  Conference on Computer Vision and Pattern Recognition}, 2017, pp. 270--279.

\bibitem{schoenberger2016mvs}
J.~L. Sch\"{o}nberger, E.~Zheng, M.~Pollefeys, and J.-M. Frahm, ``Pixelwise
  view selection for unstructured multi-view stereo,'' in \emph{European
  Conference on Computer Vision}, 2016.

\bibitem{saxena20083}
A.~Saxena, S.~H. Chung, and A.~Y. Ng, ``3-d depth reconstruction from a single
  still image,'' \emph{International journal of computer vision}, vol.~76,
  no.~1, pp. 53--69, 2008.

\bibitem{liu2010single}
B.~Liu, S.~Gould, and D.~Koller, ``Single image depth estimation from predicted
  semantic labels,'' in \emph{2010 IEEE Computer Society Conference on Computer
  Vision and Pattern Recognition}.\hskip 1em plus 0.5em minus 0.4em\relax IEEE,
  2010.

\bibitem{ladicky2014pulling}
L.~Ladicky, J.~Shi, and M.~Pollefeys, ``Pulling things out of perspective,'' in
  \emph{Proceedings of the IEEE Conference on Computer Vision and Pattern
  Recognition}, 2014, pp. 89--96.

\bibitem{liu2014discrete}
M.~Liu, M.~Salzmann, and X.~He, ``Discrete-continuous depth estimation from a
  single image,'' in \emph{IEEE Conference on Computer Vision and Pattern
  Recognition}.\hskip 1em plus 0.5em minus 0.4em\relax IEEE, 2014.

\bibitem{karsch2014depth}
K.~Karsch, C.~Liu, and S.~B. Kang, ``Depth transfer: Depth extraction from
  video using non-parametric sampling,'' \emph{IEEE transactions on pattern
  analysis and machine intelligence}, vol.~36, no.~11, pp. 2144--2158, 2014.

\bibitem{jiang2019high}
H.~Jiang and R.~Huang, ``High quality monocular depth estimation via a
  multi-scale network and a detail-preserving objective,'' in \emph{IEEE
  International Conference on Image Processing}, 2019.

\bibitem{silberman2012indoor}
N.~Silberman, D.~Hoiem, P.~Kohli, and R.~Fergus, ``Indoor segmentation and
  support inference from rgbd images,'' in \emph{European Conference on
  Computer Vision}, 2012.

\bibitem{geiger2013vision}
A.~Geiger, P.~Lenz, C.~Stiller, and R.~Urtasun, ``Vision meets robotics: The
  kitti dataset,'' \emph{The International Journal of Robotics Research},
  vol.~32, no.~11, pp. 1231--1237, 2013.

\bibitem{garg2016unsupervised}
R.~Garg, V.~K. BG, G.~Carneiro, and I.~Reid, ``Unsupervised cnn for single view
  depth estimation: Geometry to the rescue,'' in \emph{European Conference on
  Computer Vision}, 2016.

\bibitem{jaderberg2015spatial}
M.~Jaderberg, K.~Simonyan, A.~Zisserman \emph{et~al.}, ``Spatial transformer
  networks,'' in \emph{Advances in neural information processing systems},
  2015, pp. 2017--2025.

\bibitem{wang2004image}
Z.~Wang, A.~C. Bovik, H.~R. Sheikh, E.~P. Simoncelli \emph{et~al.}, ``Image
  quality assessment: from error visibility to structural similarity,''
  \emph{IEEE transactions on image processing}, vol.~13, no.~4, pp. 600--612,
  2004.

\bibitem{heise2013pm}
P.~Heise, S.~Klose, B.~Jensen, and A.~Knoll, ``Pm-huber: Patchmatch with huber
  regularization for stereo matching,'' in \emph{Proceedings of the IEEE
  International Conference on Computer Vision}, 2013.

\bibitem{tian2021depth}
F.~Tian, Y.~Gao, Z.~Fang, Y.~Fang, J.~Gu, H.~Fujita, and J.-N. Hwang, ``Depth
  estimation using a self-supervised network based on cross-layer feature
  fusion and the quadtree constraint,'' \emph{IEEE Transactions on Circuits and
  Systems for Video Technology}, 2021.

\bibitem{li2018undeepvo}
R.~Li, S.~Wang, Z.~Long, and D.~Gu, ``Undeepvo: Monocular visual odometry
  through unsupervised deep learning,'' in \emph{2018 IEEE International
  Conference on Robotics and Automation}.\hskip 1em plus 0.5em minus
  0.4em\relax IEEE, 2018.

\bibitem{zhan2018unsupervised}
H.~Zhan, R.~Garg, C.~Saroj~Weerasekera, K.~Li, H.~Agarwal, and I.~Reid,
  ``Unsupervised learning of monocular depth estimation and visual odometry
  with deep feature reconstruction,'' in \emph{Proceedings of the IEEE
  Conference on Computer Vision and Pattern Recognition}, 2018.

\bibitem{yang2018every}
Z.~Yang, P.~Wang, Y.~Wang, W.~Xu, and R.~Nevatia, ``Every pixel counts:
  Unsupervised geometry learning with holistic 3d motion understanding,'' in
  \emph{Proceedings of the European Conference on Computer Vision}, 2018.

\bibitem{watson2019self}
J.~Watson, M.~Firman, G.~J. Brostow, and D.~Turmukhambetov, ``Self-supervised
  monocular depth hints,'' in \emph{Proceedings of the IEEE International
  Conference on Computer Vision}, 2019.

\bibitem{guizilini20203d}
V.~Guizilini, R.~Ambrus, S.~Pillai, A.~Raventos, and A.~Gaidon, ``3d packing
  for self-supervised monocular depth estimation,'' in \emph{Proceedings of the
  IEEE/CVF Conference on Computer Vision and Pattern Recognition}, 2020, pp.
  2485--2494.

\bibitem{mahjourian2018unsupervised}
R.~Mahjourian, M.~Wicke, and A.~Angelova, ``Unsupervised learning of depth and
  ego-motion from monocular video using 3d geometric constraints,'' in
  \emph{Proceedings of the IEEE Conference on Computer Vision and Pattern
  Recognition}, 2018.

\bibitem{kuznietsov2017semi}
Y.~Kuznietsov, J.~St{\"u}ckler, and B.~Leibe, ``Semi-supervised deep learning
  for monocular depth map prediction,'' in \emph{The IEEE Conference on
  Computer Vision and Pattern Recognition}, 2017.

\bibitem{pillai2019superdepth}
S.~Pillai, R.~Ambru{\c{s}}, and A.~Gaidon, ``Superdepth: Self-supervised,
  super-resolved monocular depth estimation,'' in \emph{2019 International
  Conference on Robotics and Automation}.\hskip 1em plus 0.5em minus
  0.4em\relax IEEE, 2019.

\bibitem{wang2018learning}
C.~Wang, J.~Miguel~Buenaposada, R.~Zhu, and S.~Lucey, ``Learning depth from
  monocular videos using direct methods,'' in \emph{Proceedings of the IEEE
  Conference on Computer Vision and Pattern Recognition}, 2018.

\bibitem{he2016deep}
K.~He, X.~Zhang, S.~Ren, and J.~Sun, ``Deep residual learning for image
  recognition,'' in \emph{Proceedings of the IEEE conference on computer vision
  and pattern recognition}, 2016.

\bibitem{deng2009imagenet}
J.~Deng, W.~Dong, R.~Socher, L.-J. Li, K.~Li, and L.~Fei-Fei, ``Imagenet: A
  large-scale hierarchical image database,'' in \emph{IEEE conference on
  computer vision and pattern recognition}, 2009.

\bibitem{kingma2014adam}
D.~P. Kingma and J.~Ba, ``Adam: A method for stochastic optimization,''
  \emph{arXiv preprint arXiv:1412.6980}, 2014.

\bibitem{zhang2018deep}
Z.~Zhang, C.~Xu, J.~Yang, Y.~Tai, and L.~Chen, ``Deep hierarchical guidance and
  regularization learning for end-to-end depth estimation,'' \emph{Pattern
  Recognition}, pp. 430--442, 2018.

\bibitem{jiang2019hierarchical}
H.~Jiang and R.~Huang, ``Hierarchical binary classification for monocular depth
  estimation,'' in \emph{IEEE International Conference on Robotics and
  Biomimetics}, 2019.

\bibitem{kong2018pixel}
S.~Kong and C.~Fowlkes, ``Pixel-wise attentional gating for parsimonious pixel
  labeling,'' in \emph{arxiv 1805.01556}, 2018.

\bibitem{li2018deep}
R.~Li, K.~Xian, C.~Shen, Z.~Cao, H.~Lu, and L.~Hang, ``Deep attention-based
  classification network for robust depth prediction,'' in \emph{Asian
  Conference on Computer Vision}, 2018.

\bibitem{goldman2019lsim}
M.~Goldman, T.~Hassner, and S.~Avidan, ``Learn stereo, infer mono: Siamese
  networks for self-supervised, monocular, depth estimation,'' in
  \emph{Computer Vision and Pattern Recognition Workshops}, 2019.

\bibitem{labelme2016}
K.~Wada, ``{labelme: Image Polygonal Annotation with Python},''
  \url{https://github.com/wkentaro/labelme}, 2016.

\bibitem{hirschmuller2007stereo}
H.~Hirschmuller, ``Stereo processing by semi-global matching and mutual
  information,'' \emph{IEEE Transactions on pattern analysis and machine
  intelligence}, vol.~30, no.~2, pp. 328--341, 2007.

\end{thebibliography}
}



%\vfill


\end{document}